# Mechanisms for Automated Negotiation
# in State Oriented Domains

**Gilad Zlotkin**                                                 GILADZ@AGENTSOFT.COM
*AgentSoft Ltd.*
*P.O. Box 53047*
*Jerusalem, Israel*

**Jeffrey S. Rosenschein**                                       JEFF@CS.HUJI.AC.IL
*Institute of Computer Science*
*Hebrew University*
*Givat Ram, Jerusalem, Israel*

## Abstract

This paper lays part of the groundwork for a domain theory of negotiation, that is, a way of classifying interactions so that it is clear, given a domain, which negotiation mechanisms and strategies are appropriate. We define State Oriented Domains, a general category of interaction. Necessary and sufficient conditions for cooperation are outlined. We use the notion of *worth* in an altered definition of utility, thus enabling agreements in a wider class of joint-goal reachable situations. An approach is offered for conflict resolution, and it is shown that even in a conflict situation, partial cooperative steps can be taken by interacting agents (that is, agents in fundamental conflict might still agree to cooperate up to a certain point).

A Unified Negotiation Protocol (UNP) is developed that can be used in all types of encounters. It is shown that in certain borderline cooperative situations, a partial cooperative agreement (i.e., one that does not achieve all agents' goals) might be preferred by all agents, even though there exists a rational agreement that would achieve all their goals.

Finally, we analyze cases where agents have incomplete information on the goals and worth of other agents. First we consider the case where agents' goals are private information, and we analyze what goal declaration strategies the agents might adopt to increase their utility. Then, we consider the situation where the agents' goals (and therefore stand-alone costs) are common knowledge, but the *worth* they attach to their goals is private information. We introduce two mechanisms, one "strict," the other "tolerant," and analyze their affects on the stability and efficiency of negotiation outcomes.

## 1. Introduction

Negotiation has been a major research topic in the distributed artificial intelligence (DAI) community (Smith, 1978; Malone, Fikes, Grant, & Howard, 1988; Kuwabara & Lesser, 1989; Conry, Meyer, & Lesser, 1988; Kreifelts & von Martial, 1991). The term negotiation, however, has been used in a variety of different ways. To some researchers, negotiation serves as an important mechanism for assigning tasks to agents, for resource allocation, and for deciding which problem-solving tasks to undertake. In these systems, there is generally some notion of global utility that the system is trying to maximize.

Other researchers have focused on negotiation that might take place among agents that serve the interests of truly distinct parties (Rosenschein & Genesereth, 1985; Sycara, 1988; Kraus & Wilkenfeld, 1990; Zlotkin & Rosenschein, 1989). The agents are autonomous in the sense that they have their own utility functions, and no global notion of utility (not even an implicit one) plays a role in their design. Negotiation can be used to share the work associated with carrying out a joint plan, or to resolve outright conflict arising from limited resources.

Despite the varied use of terminology, it is clear to the DAI community as a whole that the operation of interacting agents would be enhanced if they were able to exchange information to reach mutually beneficial agreements.

The work described in this paper follows the general direction of previous research by the authors (Rosenschein & Genesereth, 1985; Zlotkin & Rosenschein, 1989) in treating negotiation in the spirit of game theory. The focus of this research is to analyze the existence and properties of certain kinds of deals and protocols among agents. We are not here examining the computational issues that arise in discovering such deals, though the design of efficient, possibly domain-specific, algorithms will constitute an important future phase of this research. Initial work in building a domain theory of negotiation was previously undertaken (Zlotkin & Rosenschein, 1993a), and is expanded and generalized in the current paper. This analysis serves as a critical step in applying the theory of negotiation to real-world applications.

## 1.1 Applying Game Theory Tools to Protocol Design for Automated Agents

Our ongoing research has been motivated by one, focused premise: the problem of how to get computers to interact effectively in heterogeneous systems can be tackled through the use of game theory tools.

Our concern is with computer systems made up of machines that have been programmed by different entities to pursue differing goals. One approach for achieving coordination under these circumstances is to establish mutually accepted protocols for the machines to use in coming to agreements.

The perspective of our research is that one can use game theory tools to design and evaluate these high-level protocols. We do not intend, with this paper, to make contributions to game theory itself. We are not defining new notions of equilibria, nor are we providing new mathematical tools to be used in general game theory. What we are doing is taking the game theory approach, and some of its tools, to solve specific problems of high-level protocol design.

While game theory makes contributions to the understanding of many different fields, there is a particularly serendipitous match between game theory and heterogeneous computer systems. Computers, being pre-programmed in their behavior, make concrete the notion of "strategy" that plays such a central role in game theory—the idea that a player adopts rules of behavior before starting to play a given game, and that these rules entirely control his responses during the game. This idealized player is an imperfect model of human behavior, but one that is quite appropriate for computers.

While we are not the first to apply game theoretic ideas to computer science, we are using the tools in a new way. While others have used game theory to answer the question,

"How should one program a computer to act in a given specific interaction?" we are addressing the question of how to design the rules of interaction themselves for automated agents. The approach taken in this paper is, therefore, strongly based on previous work in game theory, primarily on what is known as "Nash's Bargaining Problem" (Nash, 1950; Luce & Raiffa, 1957) or "Nash's Model of Bargaining" (Roth, 1979), "mechanism design" or "implementation theory" (Binmore, 1992; Fudenberg & Tirole, 1992), and "correlated equilibrium theory" (Aumann, 1974, 1987; Myerson, 1991; Forges, 1993). A short overview of game theory results that are used or referred to in this paper can be found in Section 9.1.

## 1.2 Overview of the Paper

In previous work, we began laying the groundwork for a domain theory of negotiation, that is, a way of classifying interactions so that it is clear, given a domain, which negotiation mechanisms and strategies are appropriate. Previously, we considered Task Oriented Domains (Zlotkin & Rosenschein, 1989, 1993a), a restricted category of interactions. In this paper, we define State Oriented Domains, a more general category of interaction.

In Section 4.4 we examine scenarios where interacting agents in State Oriented Domains can find themselves in cooperative, compromise, and conflict encounters. In conflict situations, the agents' goals cannot be simultaneously achieved. A joint-goal reachable situation (i.e., where agents' goals *can* be simultaneously achieved) can be cooperative or compromise, depending on the cost of reaching a state that satisfies all agents compared to the cost of each agent (alone) achieving his stand-alone goal.

In Section 4.1, necessary and sufficient conditions for cooperation are outlined. Cooperative situations lend themselves to mixed-joint-plan-based negotiation mechanisms. However, compromise situations require special treatment. We propose using the notion of *worth* in an altered definition of utility, thereby enabling agreements in a wider class of joint-goal reachable situations. An approach is offered for conflict resolution, and it is shown that even in a conflict situation, partial cooperative steps can be taken by interacting agents (that is, agents in fundamental conflict might still agree to cooperate up to a certain point).

A Unified Negotiation Protocol (UNP) is developed in Section 5.4 that can be used in all types of encounters. It is shown that in certain borderline cooperative situations, a partial cooperative agreement (i.e., one that does not achieve all agents' goals) might be preferred by all agents, even though there exists a rational agreement that would achieve all their goals.

The UNP is further enhanced in Section 6 to deal with the case where agents have assigned unlimited worth to their goals and this fact is common knowledge. Our solution depends on the concept of "cleaning up after yourself," or *tidiness*, as a new method of evaluating agent utility. We show that two tidy agents are able to reach agreements in *all* joint-goal reachable situations in State Oriented Domains.

In Section 7 we analyze cases where agents have incomplete information about the goals and worth of other agents. First, we consider the case where agents' goals are private information, and we consider what goal declaration strategies the agents might adopt to increase their utility.

We then consider, in Section 8, the situation where the agents' goals (and therefore stand-alone costs) are common knowledge, but the *worth* they attach to their goals is private information. There are many situations where an agent's goals might be known, but his worth is private. For example, two cars approaching an intersection may know each other's goals (because of the lanes in which they are located). The worth that each associates with passing through the intersection to his target lane, however, is private. Goal recognition techniques are suitable for discovering the other agent's intentions; his worth, however, is harder to discern from short-term external evidence.

The agents declare, in a $-1$-phase, their worths, which are then used as a baseline to the utility calculation (and thus affect the negotiation outcome). We are concerned with analyzing what worth declaration strategies the agents might adopt to increase their utility. We introduce two mechanisms, one "strict," the other "tolerant," and analyze their affects on the stability and efficiency of negotiation outcomes. The strict mechanism turns out to be more stable, while the tolerant mechanism is more efficient.

## 2. Negotiation in State Oriented Domains

How can machines decide how to share resources, or which machine will give way while the other proceeds? Negotiation and compromise are necessary, but how do we build our machines to do these things? How can the designers of these separate machines decide on techniques for agreement that enable mutually beneficial behavior? What techniques are appropriate? Can we make definite statements about the techniques' properties?

The way we address these questions is to synthesize ideas from artificial intelligence with the tools of game theory. Assuming that automated agents, built by separate, self-interested designers, will interact, we are interested in designing *protocols* for specific domains that will get those agents to interact in useful ways.

The word "protocol" means different things to different people. When we use the word protocol, we mean the rules by which agents will come to agreements. It specifies the kinds of deals they can make, as well as the sequence of offers and counter-offers that are allowed. Protocols are intimately connected with *domains*, by which we mean the environment in which our agents operate. Automated agents who control telecommunications networks are operating in a different domain (in a formal sense) than robots moving boxes. Much of our research is focused on the relationship between different kinds of domains, and the protocols that are suitable for each.

Given a protocol, we need to consider what agent *strategy* is appropriate. A strategy is the way an agent behaves in an interaction. The protocol specifies the rules of the interaction, but the exact deals that an agent proposes are a result of the strategy that his designer has put into him. As an analogy, a protocol is like the rules governing movement of pieces in the game of chess. A strategy is the way in which a chess player decides on his next move.

### 2.1 Attributes of Standards

What are the attributes that might interest protocol designers? The set of attributes, and their relative importance, will ultimately affect their choice of interaction rules.

We have considered several attributes that might be important to system designers.

1. **Efficiency:** The agents should not squander resources when they come to an agreement; there should not be wasted utility when an agreement is reached. For example, it makes sense for the agreements to satisfy the requirement of Pareto Optimality (no agent could derive more from a different agreement, without some other agent deriving less from that alternate agreement). Another consideration might be Global Optimality, which is achieved when the sum of the agents' benefits are maximized. Global Optimality implies Pareto Optimality, but not vice versa. Since we are speaking about self-motivated agents (who care about their own utilities, not the sum of system-wide utilities—no agent in general would be willing to accept lower utility just to increase the system's sum), Pareto Optimality plays a primary role in our efficiency evaluation. Among Pareto Optimal solutions, however, we might also consider as a secondary criterion those solutions that increase the sum of system-wide utilities.

2. **Stability:** No agent should have an incentive to deviate from agreed-upon strategies. The strategy that agents adopt can be proposed as part of the interaction environment design. Once these strategies have been proposed, however, we do not want individual designers (e.g., companies) to have an incentive to go back and build their agents with different, manipulative, strategies.

3. **Simplicity:** It will be desirable for the overall interaction environment to make low computational demands on the agents, and to require little communication overhead. This is related both to efficiency and to stability: if the interaction mechanism is simple, it increases efficiency of the system, with fewer resources used up in carrying out the negotiation itself. Similarly, with stable mechanisms, few resources need to be spent on outguessing your opponent, or trying to discover his optimal choices. The optimal behavior has been publicly revealed, and there is nothing better to do than just carry it out.

4. **Distribution:** Preferably, the interaction rules will not require a central decision maker, for all the obvious reasons. We do not want our distributed system to have a performance bottleneck, nor collapse due to the single failure of a special node.

5. **Symmetry:** We may not want agents to play different roles in the interaction scenario. This simplifies the overall mechanism, and removes the question of which agent will play which role when an interaction gets under way.

These attributes need not be universally accepted. In fact, there will sometimes be trade-offs between one attribute and another (for example, efficiency and stability are sometimes in conflict with one another; see Section 8). But our protocols are designed, for specific classes of domains, so that they satisfy some or all of these attributes. Ultimately, these are the kinds of criteria that rate the acceptability of one interaction mechanism over another.

As one example, the attribute of stability assumes particular importance when we consider open systems, where new agents are constantly entering and leaving the community of interacting machines. Here, we might want to maintain stability in the face of new agents who bring with them new goals and potentially new strategies as well. If the mechanism is "self-perpetuating," in that it is not only to the benefit of society as a whole to follow the rules, but also to the benefit of each individual member, then the social behavior remains stable even when the society's members change dynamically. When the interaction rules create an environment in which a particular strategy is optimal, beneficial social behavior is resistant to outside invasion.

## 2.2 Side Effects in Encounters

Various kinds of encounters among agents, in various types of domains, are possible. In previous work (Zlotkin & Rosenschein, 1989, 1993a, 1994, 1996b) we examined Task Oriented Domains (TODs), which encompass only certain kinds of encounters among agents. *State Oriented Domains* (SODs) describe a larger class of scenarios for multiagent encounters than do TODs. In fact, as we will see below, the set of Task Oriented Domains is actually a proper subset of State Oriented Domains. Most classical domains in Artificial Intelligence have been instances of State Oriented Domains.

The main attribute of general SODs is that agents' actions can have side effects. In Task Oriented Domains, no side effects exist and in general all common resources are unrestricted. Thus, when an agent achieves his own set of tasks in a TOD it has no positive nor negative effects on the other agent whatsoever. It does not hinder the other agent from achieving his goal, and it never satisfies the other agent's goals "by accident." To enable another agent to carry out your task, such as for example in the Postmen Domain (Zlotkin & Rosenschein, 1989), it is necessary explicitly to declare the existence of the letter, and hand it over, so that it will be delivered. The absence of side effects rules out some positive and all negative interactions among agent goals. The only positive interactions that remain are those that are *explicitly* coordinated by the agents.

In general State Oriented Domains, where side effects exist, agents can unintentionally achieve one another's goals, and thus benefit from one another's actions. The flip side of side effects, however, is that negative interactions between goals can also exist. Thus, an SOD is a domain that is (unlike TODs) not necessarily cooperative, because of those action side effects. In SODs, agents have to deal with goal conflict and interference, as well as the possibility of unintended cooperation.[1]

For example, consider the Blocks World situation in Figure 1. The simplest plan to achieve On(White, Gray) has the side effect of achieving Clear(Black).

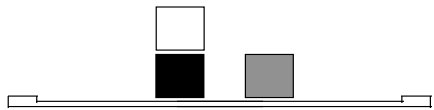

Figure 1: Side Effects in State Oriented Domains

---

1. For interesting discussions of the issue of conflict and its role in human encounters, see (Schelling, 1963, 1984).

## 2.3 Domain Definition

Consider a group of agents who co-exist in some environment. Each agent has a *goal* that it is interested in achieving. What does it mean to achieve a goal? In State Oriented Domains, it is the classic AI notion of goal achievement: it means to carry out a sequence of actions (a *plan*) that results in the transformation of the environment to a state where the goal is satisfied.

Imagine, for example, a person who is interested in getting to work. His goal is to be at work; in the current state, he is not at work. His plan will be the sequence of actions that get him to work (driving his car, or calling a taxi, or walking, or riding a bicycle,...). The final, or goal, state, may differ depending on which plan was executed (e.g., where his car is, where his bicycle is). All the states in which he is at work, however, satisfy his goal. Let's assume that the optimal plan (from the time point of view) involves driving the car to work.

The specification of the goal states may be implicit. The fact that needs to be true (the goal) may be given. Any situation in which that fact is true, i.e., the goal is satisfied, is acceptable. In a State Oriented Domain, any goal is described by the set of states that satisfy it.

Now imagine that this person's wife is interested in being at her own place of work. There are states that will satisfy both the husband's and wife's goals, and plans that will achieve such a state (e.g., one of them takes the car, while the other calls a taxi). However, there are certain plans that are suitable for either spouse in isolation, but which cannot coexist. For example, the husband taking the car is a perfectly good plan (and optimal) if he were alone in the world. Similarly, his wife's taking the car is a good plan (and optimal) if she were alone. Together, another plan may be suitable (husband drives wife to her work, continues on with car to his work). In this case, extra work was required from the husband's point of view, because the wife is present in his world; there is a certain burden to the coordination.

In the example above, the agents carry out a sequence of activities, suitably synchronized, to reach the goal state satisfying both. The husband and wife enter the car, after which the husband drives to a particular location, the wife exits, and so on. In any environment, there are primitive operations that each agent alone can do. When these operations are combined into a coherent sequence of actions specifying what *both* agents are to do (and the order in which they are to be done), we say that the agents are executing a *joint plan*. A joint plan in general transforms the world from some initial state to a goal state satisfying both agents (when possible). The plan above transforms the world from the initial state where both husband and wife are at home to the goal state (satisfying both agents) where the wife is at work, and the car and husband are at his place of work. This is the final state, one of many goal states.

In Task Oriented Domains the cost of the coordinated plan need never be worse than the stand-alone plan—at the very worst, each agent just achieves his own set of tasks. In our husband/wife sharing one car example, however, the coordinated plan may be worse for one or both agents than their stand-alone plans. This is an example of one attribute of State Oriented Domains, namely negative interactions, or what are sometimes called

"deleted-condition interactions" (Gupta & Nau, 1992). This is because taking the car has the side effect of depriving the other agent of the car.

Imagine a new situation, that arises during the weekend. The husband is interested in doing carpentry in the garage (currently occupied by the car). The wife is interested in taking the car to the baseball game. By themselves, each agent has an optimal plan to reach a goal state (e.g., the husband moves the car out of the garage, parks it outside, does his carpentry). However, when his wife takes the car to the game, executing her stand-alone optimal plan, the husband benefits from the side effect of the car being moved, namely, the garage is emptied. This is an example of another typical attribute of State Oriented Domains—accidental achievement of goals, or "enabling-condition interactions" (Gupta & Nau, 1992) or "favor relations" (von Martial, 1990) among goals.

When agents carry out a joint plan, each one plays some "role." Our theory assumes that there is some way of assessing the cost of each role. This measure of cost is essential to how an agent evaluates a given joint plan. Among all joint plans that achieve his goal, he will prefer those in which his role has lower cost.

We express the intuitive ideas above in the precise definition below.

**Definition 1** *A State Oriented Domain (SOD) is a tuple $< \mathcal{S}, \mathcal{A}, \mathcal{J}, c >$ where:*

1. *$\mathcal{S}$ is the set of all possible world states;*

2. *$\mathcal{A} = \{A_1, A_2, \ldots A_n\}$ is an ordered list of agents;*

3. *$\mathcal{J}$ is the set of all possible joint (i.e., n-agent) plans. A joint plan $J \in \mathcal{J}$ moves the world from one state in $\mathcal{S}$ to another. The actions taken by agent $k$ are called $k$'s role in $J$, and will be written as $J_k$. We can also write $J$ as $(J_1, J_2, \ldots, J_n)$;*

4. *$c$ is a function $c: \mathcal{J} \to (\mathbb{R}^+)^n$. For each joint plan $J$ in $\mathcal{J}$, $c(J)$ is a vector of $n$ positive real numbers, the cost of each agent's role in the joint plan. $c(J)_i$ is the $i$-th element of the cost vector, i.e., it is the cost of the $i$-th role in $J$. If an agent plays no role in $J$, his cost is $0$.*

Our use of the term *joint plan* differs from other uses in the AI literature (Levesque & Cohen, 1990; Cohen & Levesque, 1991). There, the term joint plan implies a joint *goal*, and mutual commitment by the agents to full implementation of the plan (e.g., if one agent dropped out suddenly, the other would still continue). In our use of the term, the agents are only committed to their own goal and their *part* of the combined plan. Each may do its part of the plan for different reasons, because each has a different goal to achieve. Were one agent to drop out, the other agent may or may not continue, depending on whether it suited his own goal.

The details of the description of the joint plans in $\mathcal{J}$ are not critical to our overall theory. The minimal requirement is that it must be possible to evaluate the cost of the joint plan for each agent (i.e., the cost of his role). In many domains, a joint plan will be a sequence of actions for each agent with an associated schedule (partial order) constraining the actions' parallel execution.

Note also that our cost function above relates only to the joint plan itself and not, for example, to the initial state of the world. In fact, the cost function could be altered to include other parameters (like the initial state of the world), without affecting our discussion below. Our model is not sensitive to the details of the cost function definition, other than the requirement that the cost of a role be the same for all agents. This is called the *symmetric abilities* assumption (see below, Section 2.4).

**Definition 2** *An* encounter *within an SOD* $< \mathcal{S}, \mathcal{A}, \mathcal{J}, c >$ *is a tuple* $< s, (G_1, G_2, \ldots, G_n) >$ *such that* $s \in \mathcal{S}$ *is the initial state of the world, and for all* $k \in \{1 \ldots n\}$, $G_k$ *is the set of all acceptable final world states from* $\mathcal{S}$ *for agent* $A_k$. $G_k$ *will also be called* $A_k$'s goal.

An agent's goal is a fixed, pre-determined, set of states. An agent will, at the conclusion of the joint plan, either achieve his goal or not achieve his goal. Goals cannot be partially achieved. Domains in which goals *can* be partially achieved are called Worth Oriented Domains (WODs) and are discussed in detail elsewhere (Zlotkin & Rosenschein, 1991c, 1996a).

One thing that we are specifically ruling out in SODs is one agent having a goal that makes reference to another agent's (as yet) unknown goal. For example, a specification such as "Agent 1's goal is to make sure that Agent 2's goal will not be achieved, whatever the latter's goal is" cannot constitute part of the description of an encounter in a State Oriented Domain, because it cannot be described as a static set of goal states. However, this *meta-goal* might exist within an agent, and give rise to a well-defined set of states in a specific encounter (e.g., given $G_2$, $G_1$ is its complement). Similarly, one agent might have as its goal that another agent have a specific goal $G_2$—the first agent wants the world to be in a state where the other agent has the specific goal $G_2$.

We will only consider sets of goal states that can be specified in a finite way, either because the set itself is finite, or the infinite set can be specified by a closed formula in first-order logic (i.e., no free variables; all states that satisfy the formula, and only those states, are in the goal set). As an example, an agent might have the goal that "There exists a block $x$ such that block B is on $x$."

We will also consider further restrictions on the kind of goals agents may have. For example, below we will consider domains in which agents' goals are restricted to sets of grounded predicates (i.e., no variables) rather than to any closed formula.

### 2.3.1 REACHABILITY

It may be the case that there exist goal states that satisfy both agents' goals, but that there are constraints as to the reachability of those states. For example, it may be the case that a state satisfying each goal can be reached by an agent alone, but a state satisfying the combined goal cannot be reached by any agent alone. More generally, reaching any state might require $n$ agents working together, and be unreachable if fewer than $n$ agents are involved (we will call $n$ the "parallelism factor" of the goal). When the goal in the intersection cannot be reached by any number of agents working in parallel, we will say the parallelism factor is infinite. The parallelism factor is a particularly appropriate concept when there are multiagent actions that are possible or required in the domain (e.g., carrying a heavy table).

## 2.4 Assumptions

Throughout this paper, we will be making a number of simplifying assumptions that enable us to lay out the foundation for our theory of mechanism design for automated agents. Here, we present those assumptions.

1. **Expected Utility Maximizer:** Designers will design their agents to maximize expected utility. For example, we assume that a designer will build his agent to prefer a 51% chance of getting $100, rather than a sure $50.

2. **Isolated Negotiation:** An agent cannot commit himself as part of the current negotiation to some behavior in a future negotiation, nor can he expect that his current behavior will in any way affect a future negotiation. Similarly, an agent cannot expect others to behave in a particular way based on their previous interaction history, nor to act differently with him because of his own past behavior. Each negotiation stands alone.

3. **Interagent Comparison of Utility:** The designers have a means of transforming the utilities held by different agents into common utility units.

4. **Symmetric Abilities:** All agents are able to perform the same set of operations in the world, and the cost of each operation is independent of the agent carrying it out.

5. **Binding Commitments:** Designers will design their agents to keep explicit public commitments. We assume nothing about the relationship between private preferences and public behavior, only that public commitment be followed by public performance of the commitment. This can be monitored, and if necessary, enforced.

6. **No Explicit Utility Transfer:** Although agents can compare their respective utilities, they have no way of explicitly transferring utility units from one to the other. There is, for example, no "money" that can be used to compensate one agent for a disadvantageous agreement. Utility transfer does occur, however, implicitly. This implicit transfer of utility forms the basis for agreement among agents.

## 3. Examples of State Oriented Domains

In this section, we present several examples of State Oriented Domains. These specific examples illustrate some of the nuances of describing this class of domains.

## 3.1 The Blocks Domain

In the Blocks Domain, there is a table of unlimited size, and a set of blocks. A block can be on the table or on some other block, and there is no limit to the height of a stack of blocks. One state in this domain can be seen in Figure 2.

*World States and Goals:* The basic predicates that make up world states and goals are:

- On$(x, y)$: such that $x$ and $y$ are blocks; its meaning is that block $x$ is (directly) on block $y$.

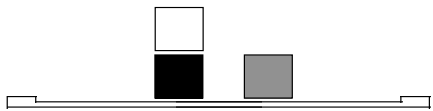

Figure 2: A State in the Blocks Domain

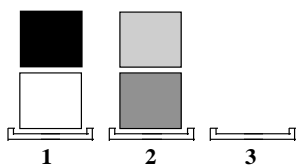

Figure 3: A State in the Slotted Blocks Domain

- On($x$, *Table*): such that $x$ is a block; its meaning is that block $x$ is (directly) on the table.

- Clear($x$): such that $x$ is a block; its meaning is that there is no block on $x$, i.e., Clear($x$) ≡ ¬∃$y$ On($y$, $x$).

As this is an SOD, goals are sets of world states. These world states can be expressed as a first order closed formula over the above predicates. Sample goals are:

- ¬Clear($R$) — Block $R$ is not clear.

- ∃$x$ On($R$, $x$) — Block $R$ is not on the table.

- ∀$x$ On($x$, Table) — All blocks are on the table (and therefore, implicitly, all blocks are also Clear).

*Atomic Operation:* There is one operation in this world: Move($x$, $y$). This operation moves a clear block $x$ onto the top of another clear block $y$.

*Cost:* Each move operation has a cost of 2.

## 3.2 The Slotted Blocks Domain

The domain here is the same as the Blocks Domain above. However, on the table there are only a bounded number of slots into which blocks can be placed. One state in this domain can be seen in Figure 3.

*World States and Goals:* The basic predicates that make up world states and goals are:

- On($x$, $y$): such that $x$ and $y$ are blocks; its meaning is that block $x$ is (directly) on block $y$.

- At$(x, n)$: such that $x$ is a block and $n$ is a slot name; its meaning is that block $x$ is (directly) on the table at slot $n$.

- Clear$(x)$: such that $x$ is a block; its meaning is that there is no block which is on $x$, i.e., Clear$(x) \equiv \neg \exists y$ On$(y, x)$.

*Atomic Operations:* There are two operations in the Slotted Blocks Domain:

- PickUp$(i)$ — Pick up the top block in slot $i$ (can be executed whenever slot $i$ is not empty);

- PutDown$(i)$ — Put down the block that is currently being held into slot $i$. An agent can hold no more than one block at a time.

*Cost:* Each operation has a cost of one.

This Slotted Blocks Domain is different from the Blocks Domain above in two ways:

1. The table of unlimited size is replaced by a bounded table with distinguishable locations that we call "slots."

2. The atomic "Move" operation is broken into two sub-operations PickUp and PutDown. This allows more cooperation among the agents. For example, if we want to swap the blocks in slot 1 in Figure 3 it would take one or more agents minimally a total of 4 Move operations, i.e., each block (Black and White) is touched twice. However, if we allow the agents to use the PickUp and PutDown operations two agents can do the swap with two PickUp and two PutDown operations (which is equivalent to two move operations), i.e., each block is touched only once. The finer granularity of the operations allows more flexibility in scheduling within the joint plan.

### 3.3 The Delivery Domain with Bounded Storage Space

In this Delivery Domain, there is a weighted graph $G = G(V, E)$. Each $v \in V$ represents a warehouse, and each $e \in E$ represents a road. The weight function $w: E \to \mathbb{R}^+$ is the length of any given road. For each edge $e \in E$, $w(e)$ is the length of $e$ or the "cost" of $e$. Each agent has to deliver containers from one warehouse to another. To do the deliveries, agents can rent trucks, an unlimited supply of which are available for rental at every node. A truck can carry up to 5 containers. Each warehouse also has a limited capacity for holding containers.

*Atomic Operations:* The operations in this domain are:

- Load$(c, t)$ — loads a container $c$ onto a truck $t$. The preconditions are:
  - Container $c$ and truck $t$ are at the same warehouse $h$;
  - Truck $t$ has less than 5 containers on board, where 5 is the capacity limit of each truck.

  The results of the operation are:
  - Warehouse $h$ has one container less;

– Truck $t$ has one container more.

A Load operation costs 1.

- Unload($c, t$) — unloads a container $c$ from a truck $t$. The preconditions are:

  – Container $c$ is on truck $t$;

  – Truck $t$ is at some warehouse $h$;

  – Warehouse $h$ is not full.

  The results of the operation are:

  – Warehouse $h$ has one container more;

  – Truck $t$ has one container less.

  The Unload operation costs 1.

- Drive($t, h$) — Truck $t$ drives to warehouse $h$. There are no preconditions on this operation. The result is that truck $t$ is at warehouse $h$. The cost of this operation is equal to the distance (i.e., the minimal weighted path) between the current position of truck $t$ and warehouse $h$.

*World States and Goals:* The full description of a world state includes the location of each container (either in some warehouse or on some truck) and the location of each truck (either in some warehouse or on some road). However, we will restrict goals so that they can only specify which containers need to be at which warehouses.

### 3.4 The Restricted Usage Shared Resource Domain

In this domain, there is a set of agents that are able to use a shared resource (such as a communication line, a shared memory device, a road, a bridge...). There is a restriction that no more than $m \geq 1$ agents can use the resource at the same time ($m$ denotes the maximal capacity of the resource).

*Atomic Operations:* The atomic operations in the Shared Resource Domain are:

- Use — an agent is using the shared resource for one time unit. The Use operation costs 0.

- Wait — an agent is waiting to use the shared resource for one time unit. The operation costs 1, i.e., waiting for one time unit to access the shared resource costs 1.

- $NOP$ — an agent does not need the resource and therefore neither uses it nor waits for it. This operation costs 0.

The objective here is to find a schedule such that at any time unit no more than $m$ agents are performing the Use operation.

*World States and Goals:* A world state describes the current activity of the agents and their accumulated *resource usage* since time 0 (i.e., not their accumulated cost). The goal of an agent is to be in a state where it has accumulated a target number of time units using the resource, and is currently doing the NOP operation. Formally, a state is an $n$-element vector, one element for each agent, where each element is a pair consisting of the agent's current operation and his accumulated number of time units using the resource (i.e., the set of all states is $(\{\text{Wait},\text{Use},\text{NOP}\} \times \mathbb{N})^n$).

|   |   | **Joint Plan** | | | World States | | |
|---|---|---|---|---|---|---|---|
|   |   | $A_1$ | $A_2$ | $A_3$ | $A_1$ | $A_2$ | $A_3$ |
| **T** | 0 | Use | Use | Wait | (Use,0) | (Use,0) | (Wait,0) |
| **i** | 1 | Use | Use | Wait | (Use,1) | (Use,1) | (Wait,0) |
| **m** | 2 | NOP | Use | Use | (NOP,2) | (Use,2) | (Use,0) |
| **e** | 3 | NOP | NOP | Use | (NOP,2) | (NOP,3) | (Use,1) |
|   | 4 | NOP | NOP | NOP | (NOP,2) | (NOP,3) | (NOP,2) |

Figure 4: Joint Plan and States in the Restricted Usage Shared Resource Domain

Assume, as an example, that there are three agents, and one resource that has a maximal capacity of two. Agents 1 and 3 need two units of the resource, while agent 2 needs three units of the resource. A joint plan can be seen at the left side of Figure 4, described by a matrix. For each time $t$ and agent $A_i$, the entry in column $i$ and row $t$ is agent $A_i$'s action at time $t$. The resulting world state after each time unit of the joint plan can be seen at the right side of Figure 4. The final state satisfies all agents' goals.

## 4. Deals, Utility, and Negotiation Mechanisms

Now that we have defined the characteristics of a State Oriented Domain, and looked at a few simple examples, we turn our attention to how agents in an SOD can reach agreement on a joint plan that brings them to some agreed-upon final state. Hopefully, this final state will satisfy both agents' goals. However, this isn't always possible. There are three such cases:

1. It might be the case that there doesn't exist a state that satisfies both agents' goals (i.e., the goals contradict one another);

2. It might be the case that there exists a state that satisfies them both, but it cannot be reached with the primitive operations in the domain (see Section 2.3.1 above);

3. It might be the case that there exists a reachable state that satisfies them both, but which is so expensive to get to that the agents are unwilling to expend the required effort.

### 4.1 A Negotiation Mechanism

We will start by presenting a simple mechanism that is suitable for cases where there exists a reachable final state (that is, reachable by a sufficiently inexpensive plan) that satisfies both agents' goals. We call this a *cooperative situation*. Later, we will enhance the mechanism so that it can handle all possible encounters in State Oriented Domains, i.e., so that it can handle conflict resolution.

**Definition 3** *Given an SOD $< \mathcal{S}, \mathcal{A}, \mathcal{J}, c >$, we define:*

- *$\mathcal{P} \subset \mathcal{J}$ to be the set of all one-agent plans, i.e., all joint plans in which only one agent has an active role.*

- *The cost $c(P)$ of a one-agent plan in which agent $k$ has the active role, $P \in \mathcal{P}$, is a vector that has at most one non-zero element, in position $k$. When there is no likelihood of confusion, we will use $c(P)$ to stand for the $k$-th element (i.e., for $c(P)_k$), rather than the entire vector.*

**Definition 4 Best Plans**

- *$s \xrightarrow{k} f$ is the minimal cost one-agent plan in $\mathcal{P}$ in which agent $k$ plays the active role and moves the world from state $s$ to state $f$ in $\mathcal{S}$.*

- *If a plan like this does not exist then $s \xrightarrow{k} f$ will stand for some constant plan $\bowtie_k$ that costs infinity to agent $k$ and $0$ to all other agents.*

- *If $s = f$ then $s \xrightarrow{k} f$ will stand for the empty plan $\Lambda$ that costs $0$ for all agents.*

- *$s \xrightarrow{k} F$ (where $s$ is a world state and $F$ is a set of world states) is the minimal cost one-agent plan in $\mathcal{P}$ in which agent $k$ plays the active role and moves the world from state $s$ to one of the states in $F$:*

$$c(s \xrightarrow{k} F) = \min_{f \in F} c(s \xrightarrow{k} f).$$

As we mentioned above, for the moment we will be restricting our attention to encounters where there does exist one or more states that satisfy both agents' goals. What if more than one such state exists? Which state should the agents choose to reach? And what if there is more than one joint plan to reach those states? Which joint plan should the agents choose?

For example, let's say that there are two states that satisfy both agents' goals. State 1 has two possible roles, with one of the roles costing 6 and the other costing 3. State 2 has two roles also, with both of them costing 5. While State 1 is cheaper overall to reach, State 2 seems to allow for a fairer division of labor among the agents.

Assuming that the agents want their agreement to be efficient, they will decide to reach State 1. But which agent should do the role that costs 6, and which should do the role that costs 3? Our approach will be to allow them to agree on a "lottery" that will equalize the benefit they derive from the joint plan. Although eventually one agent will do more than the other, the expected benefit for the two agents will be identical (prior to holding the lottery). These plans that include a probabilistic component are called *mixed joint plans*.

Throughout this paper, we limit the bulk of our discussion about mechanisms to two-agent encounters. Initial work on the generalization of these techniques to encounters among more than two agents can be found elsewhere (Zlotkin & Rosenschein, 1994). That research considers issues of coalition formation in $n$-agent Task Oriented Domains.

**Definition 5 Deals** *Given an encounter in a two-agent SOD $< s, (G_1, G_2) >$:*

- *We define a* Pure Deal *to be a joint plan* $J \in \mathcal{J}$ *that moves the world from state $s$ to a state in $G_1 \cap G_2$.*

- *We define a* Deal *to be a mixed joint plan* $J : p; \ 0 \leq p \leq 1 \ \in \mathbb{R}$ *such that $J$ is a Pure Deal.*

The semantics of a Deal is that the agents will perform the joint plan $(J_1, J_2)$ with probability $p$, or the symmetric joint plan $(J_2, J_1)$ with probability $1 - p$ (where the agents have switched roles in $J$). Under the *symmetric abilities* assumption from Section 2.4, both agents are able to execute both parts of the joint plan, and the cost of each role is independent of which agent executes it.

**Definition 6**

- *If $\delta = (J : p)$ is a Deal, then* $\mathrm{Cost}_i(\delta)$ *is defined to be* $pc(J)_i + (1 - p)c(J)_k$ *(where $k$ is $i$'s opponent).*

- *If $\delta$ is a Deal, then* $\mathrm{Utility}_i(\delta)$ *is defined to be* $c(s \to G_i) - \mathrm{Cost}_i(\delta)$.

The utility (or benefit) for an agent from a deal is simply the difference between the cost of achieving his goal alone and his expected part of the deal. Note that we write (for example) $c(s \to G_i)$ rather than $c(s \xrightarrow{k} G_i)$, since the cost of the plan is independent of the agent that is executing it.

**Definition 7**

- *A Deal $\delta$ is* individual rational *if, for all $i$,* $\mathrm{Utility}_i(\delta) \geq 0$.

- *A Deal $\delta$ is* pareto optimal *if there does not exist another* Deal *that dominates it— there does not exist another* Deal *that is better for one of the agents and not worse for the other.*

- *The* negotiation set NS *is the set of all the deals that are both individual rational and pareto optimal.*

A necessary condition for the negotiation set not to be empty is that there be no contradiction between the two agents' goals, i.e., $G_1 \cap G_2 \neq \emptyset$. All the states that exist in the intersection of the agents' goal sets might, of course, not be reachable given the domain of actions that the agents have at their disposal. The condition of reachability is not sufficient for NS not to be empty, however, because even when there is no contradiction between agents' goals, there may still not be a cooperative solution for them. In such a situation, any joint plan that satisfies the union of goals will cost one agent (or both) more than he would have spent achieving his own goal in isolation (that is, no deal is individual rational).

As an example in the Slotted Blocks Domain, consider the following encounter. The initial state can be seen at the left in Figure 5. $A_1$'s goal is "The White block is at slot 2 but not on the table" and $A_2$'s goal is "The Black block is at slot 1 but not on the table."

To achieve his goal alone, each agent has to execute one PickUp and then one PutDown; $c(s \to G_i) = 2$. The two goals do not contradict one another, because there exists a state in the world that satisfies them both (where the White and Black blocks are each placed on a Gray block). There does not exist a joint plan that moves the world from the initial state to a state that satisfies the two goals with total cost less than eight[2]—that is, no deal is individual rational.

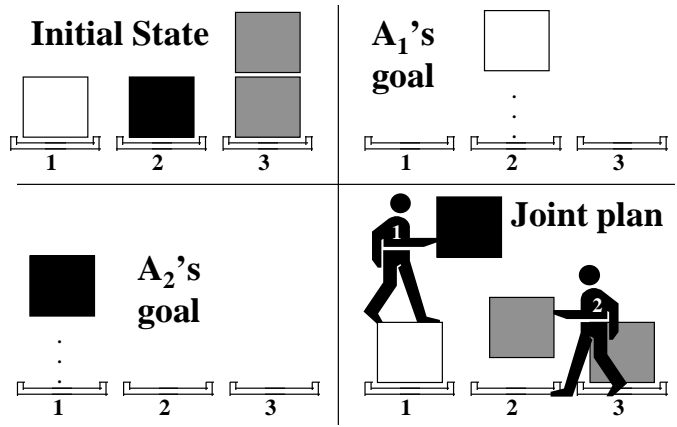

Figure 5: Conflict Exists Even Though Union of Goals is Achievable

The existence of a joint plan that moves the world from its initial state $s$ to a mutually-desired state in $G_1 \cap G_2$ is a necessary (but not sufficient) condition for the negotiation set to be non-empty. For the agents to agree on any joint plan, it should be individual rational. This means that the sum of the roles that the agents play should not exceed the sum of their individual stand-alone costs (otherwise, at least one of the agents would not get positive utility, i.e., do more work than in his stand-alone plan). But even this condition is not sufficient to guarantee an individual rational deal, since it can be the case that the minimal role in the joint plan is still too expensive for the agent with the minimum stand-alone cost. Even a probabilistic mixture of the two roles will not reduce the expected cost for that agent below the cost of the minimal role (and thus that role will not be individual rational for him).

We now show, however, that the combination of these conditions is necessary and sufficient for the negotiation set not to be empty.

**Theorem 1** *A necessary and sufficient condition for the negotiation set not to be empty is the existence of a joint plan that moves the world from its initial state $s$ to a state in $G_1 \cap G_2$ and also satisfies the following two conditions (the sum condition and the min condition):*

- *A joint plan $J$ satisfies the* sum *condition if*

$$\sum_{i=1}^{2} c(s \rightarrow G_i) \geq \sum_{i=1}^{2} c(J)_i.$$

---

2. One agent lifts the white block, while the other agent rearranges the other blocks suitably (by picking up and putting down each block once), whereupon the white block is put down. This is the best plan because each block is picked up and put down only once.

- *A joint plan $J$ satisfies the* min condition *if*

$$\min_{i=1}^{2} c(s \rightarrow G_i) \geq \min_{i=1}^{2} c(J)_i.$$

**Proof:**

- If NS $\neq \emptyset$, then let $J{:}p$ be some mixed joint plan in NS; thus, it is individual rational. $\forall i \in \{1,2\}$

$$
\begin{aligned}
\text{Utility}_i(J{:}p) &\geq 0 \\
c(s \rightarrow G_i) - \text{Cost}_i(J{:}p) &\geq 0 \\
c(s \rightarrow G_i) &\geq \text{Cost}_i(J{:}p) \\
c(s \rightarrow G_i) &\geq pc(J)_i + (1-p)c(J)_k \\
&\geq min_{l \in \{1,2\}} c(J)_l \\
\sum_{i \in \{1,2\}} c(s \rightarrow G_i) &\geq \sum_{i \in \{1,2\}} c(J)_i \\
\min_{i \in \{1,2\}} c(s \rightarrow G_i) &\geq \min_{i \in \{1,2\}} c(J)_i
\end{aligned}
$$

- Let $J$ be a *minimal total* cost joint plan that moves the world from state $s$ to a state in $G_1 \cap G_2$, and also satisfies the sum and min conditions. To show that NS $\neq \emptyset$, it is sufficient to show that there exists a deal that is both individual rational and pareto optimal. Without loss of generality, we can assume that $c(s \rightarrow G_2) \geq c(s \rightarrow G_1)$ and $c(J)_2 \geq c(J)_1$. From the min condition, we see that $c(s \rightarrow G_1) \geq c(J)_1$. There are two cases:

  - If $c(s \rightarrow G_2) \geq c(J)_2$, then the deal $J{:}1$ is individual rational.
  - If $c(s \rightarrow G_2) < c(J)_2$, then the deal $J{:}p$ (where $p = 1 - \frac{c(s \rightarrow G_1) - c(J)_1}{c(J)_2 - c(J)_1}$) is individual rational.

  $J{:}p$ is also pareto optimal, because if there is another deal $J'{:}q$ that dominates $J{:}p$ then $J'{:}q$ is also individual rational and therefore satisfies the **min** and **sum** conditions (see the proof, above, of the initial half of this theorem).

  Since $J'{:}q$ dominates $J{:}p$ it also implies that

$$\sum_{i \in \{1,2\}} \text{Utility}_i(J'{:}q) > \sum_{i \in \{1,2\}} \text{Utility}_i(J{:}p).$$

  This can be true only if

$$\sum_{i \in \{1,2\}} c(J')_i < \sum_{i \in \{1,2\}} c(J)_i.$$

  But this contradicts the fact that $J$ is the *minimal total* cost joint plan that satisfies the sum and min conditions. $\qquad \square$

The sum condition states that the sum of roles does not exceed the sum of the individual agents' stand-alone costs. The min condition states that the minimal role in the joint plan is less than the minimum stand-alone cost.

When the conditions of Theorem 1 are true, we will say that the encounters are *cooperative*. In such encounters, the agents can use some negotiation mechanism over mixed joint plans. The question we next examine is what kind of negotiation mechanism they should use.

## 4.2 Mechanisms that Maximize the Product of Utilities

In general we would like the negotiation mechanism to be symmetrically distributed, and we would also like for there to be a negotiation strategy (for that mechanism) that is in equilibrium with itself. A symmetrically distributed mechanism is one in which all agents play according to the same rules, e.g., there are no special agents that have a different responsibility in the negotiation process. When asymmetric negotiation mechanisms are used, the problem of responsibility assignment needs to be resolved first (e.g., who will be the coordinator agent). We would then need a special mechanism for the responsibility assignment negotiation. If this mechanism is also asymmetric we will need another mechanism, and so on. Therefore, it is better to have a symmetric negotiation mechanism to start with.

Among the symmetric mechanisms, we will prefer those that have a symmetric negotiation strategy that is in equilibrium. Given a negotiation mechanism $M$, we will say that a negotiation strategy $S$ from $M$ is in equilibrium if, under the assumption that all other agents are using strategy $S$ when using $M$, I (or my agent) cannot do better by using a negotiation strategy different than $S$.

Among all symmetrically distributed negotiation mechanisms that have a symmetric negotiation strategy that is in equilibrium, we will prefer those that maximize the product of agents' utilities. This means that if agents play the equilibrium strategy, they will agree on a deal that maximizes the product of their utilities. If there is more than one product-maximizing deal, they will agree on a deal (among those product maximizers) that maximizes the sum of utilities. If there is more than one sum-maximizing product maximizer, the protocol will choose among those deals with some arbitrary probability. This definition implies both individual rationality and pareto optimality of the agreed-upon deals.

Note that maximization of the product of the utilities is not a decision that agents are assumed to be making at run-time; it is a property of the negotiation mechanism agreed upon by the agent designers (i.e., we are exploring what happens when the protocol designers would agree on this property). In more classic game theory terms (see Section 9.1), the protocol acts as a kind of "mediator," recommending "maximization of product of the utilities" in all cases.

We will call this class of mechanisms the *Product Maximizing Mechanisms*, or PMMs. In previous work on TODs (Zlotkin & Rosenschein, 1989, 1993a) we presented the Monotonic Concession Protocol and the One-Step Protocol, both of which are PMMs. As mentioned above, this paper does not examine the computational issues that arise in discovering deals.

There are a number of existing approaches to the bargaining problem in game theory. One of the earliest and most popular was Nash's axiomatic approach (Nash, 1950; Luce

& Raiffa, 1957). Nash was trying to axiomatically define a "fair" solution to a bargaining situation. He listed the following criteria as ones that a fair solution would satisfy:

1. Individual rationality (it would not be fair for a participant to get less than he would anyway without an agreement);

2. Pareto Optimality (a fair solution will not specify an agreement that could be improved for one participant without harming the other);

3. Symmetry (if the situation is symmetric, i.e., both agents would get the same utility without an agreement, and for every possible deal, the symmetric deal is also possible, then a fair solution should also be symmetric, i.e., give both participants the same utility);

4. Invariance with respect to linear utility transformations. For example, imagine two agents negotiating over how to divide $100. If one agent measures his utility in dollars while the other measures his in cents, it should not influence the fair solution. Similarly, if one agent already has $10 in the bank, and evaluates the deal that gives him $x$ dollars as having utility $10 + x$ while the other evaluates such a deal as having utility $x$, it should not influence the fair solution (i.e., change of origin doesn't affect the solution);

5. Independence of irrelevant alternatives. Imagine two agents negotiating about how to divide 10,000 cents. The Nash solution will be 5,000 cents for each, due to the symmetry assumption above. Now imagine that the same agents are negotiating over $100. Even though there are now some deals that they can't reach (for example, the one where one agent gets $49.99, and the other gets $50.01), the solution should be the same, because the original solution of 5,000 cents can still be found in the new deal space.

Nash showed that the product maximizing solution not only satisfies the above criteria, but it is the only solution that satisfies them. The first four criteria above are explicitly or implicitly assumed in our own approach (in fact, for example, our version of the fourth assumption above is more restrictive than Nash's). The fifth criteria above is not *assumed* in our work, but turns out to be true in some cases anyway. We use the Nash solution, in general, as a reasonable bargaining outcome, when it is applicable. Nash, however, had some assumptions about the space of deals that we do not have. For example, the Nash bargaining problem assumes a bounded, convex, continuous, and closed region of negotiation. In our agent negotiations, we do not assume that the space of deals is convex, nor that it is continuous.

## 4.3 Worth of a Goal

When the encounter is *cooperative*, then agents can use some PMM over mixed joint plans. Such a mechanism guarantees a fair and efficient cooperative agreement. The question now, however, is what can be done in *non-cooperative* encounters?

Consider again the encounter from the Restricted Usage Shared Resource Domain where there are three agents, and one resource which has a maximal capacity of two. Agents 1

|  | Agents | | |
|---|---|---|---|
|  | $A_1$ | $A_2$ | $A_3$ |
| 0 | Use | Use | Wait |
| 1 | Use | Use | Wait |
| 2 | NOP | Use | Use |
| 3 | NOP | NOP | Use |
| 4 | NOP | NOP | NOP |

|  | Agents | | |
|---|---|---|---|
|  | $A_1$ | $A_2$ | $A_3$ |
| 0 | Use | Wait | Use |
| 1 | Use | Wait | Use |
| 2 | NOP | Use | NOP |
| 3 | NOP | Use | NOP |
| 4 | NOP | Use | NOP |
| 5 | NOP | NOP | NOP |

(Time)

Figure 6: Two Joint Plans in the Restricted Usage Shared Resource Domain

and 3 need two units of the resource while agent 2 needs 3 units of the resource. Each agent, were it alone in the world, could achieve its goal at no cost (i.e., without waiting for the resource). However, since the maximal capacity of the resource is two, the three agents together cannot achieve their combined goal without some agent having to wait. Two possible joint plans that achieve all agents' goals can be seen in Figure 6. The left joint plan gives agents 1 and 2 utility of 0, while giving agent 3 utility of −2. The right joint plan gives agents 1 and 3 utility of 0, while giving agent 2 utility of −2. Globally, the plan on the left finishes sooner. But from the perspective of the individual agents, the two plans are really comparable—in one, agent 3 suffers by waiting two time units, and in the other, agent 2 suffers by exactly the same amount. We have assumed, however, that the agents are not concerned with the global aspects of resource utilization, and are only concerned about their own local cost. In addition, both plans are Pareto Optimal, and neither of them is individual rational (because one agent gets negative utility).

If there exists a joint plan $J$ that brings the world to a state that satisfies all agents' goals, but either the min condition or the sum condition is not true, then for the agents cooperatively to bring the world to a state that satisfies all agents' goals, at least one of them will have to do more than if he were alone in the world and achieved only his own goals. Will either one of them agree to do extra work? It depends on how important each goal is to each agent $i$, i.e., how much $i$ is willing to pay to bring the world to a state in $G_i$.

For example, in the Shared Resource encounter above, agent 2 or 3 might be willing to wait two time units so as to get its turn at the resource. Although they could have done better were they alone in the world, they must cope with the presence of those other agents. With our original definition of utility, no deal that achieves all agents' goals will be individual rational—someone will have to wait, and thus get negative utility. So with that utility definition, no agent should be willing to wait. The agents will fail to reach agreement, and no one will achieve his goal. This is because utility was calculated as the difference between the cost of an agent's plan were it alone in the world and the cost of his role in the joint plan with other agents.

However, why should the agents use stand-alone cost as their baseline for determining utility? It may be the case that agents will be willing, in the presence of other agents, to admit the need to pay an extra cost, a sort of "coordination overhead." The fact that with other agents around they have to do more does not necessarily make it irrational to do more.

In Task Oriented Domains (Zlotkin & Rosenschein, 1989, 1993a, 1994), it is reasonable to use stand-alone cost as the utility baseline since there was never any coordination overhead. In the worst case, an agent could always achieve his goal at the stand-alone price, and coordination could only improve the situation. In State Oriented Domains, however, it makes sense to consider altering the utility baseline, so that agents can rationally coordinate even when there exists a coordination overhead. One way of doing this is to assume that each agent has some upper bound on the cost that he is willing to bear to achieve his goal. Then, the agent's utility can be measured relative to this upper bound. We call this upper bound the *worth* of the agent's goal.

Even in TODs, one can conceive of stand-alone cost as the worth that an agent assigns to achieving a goal. The stand-alone cost is then the maximum that an agent is willing to expend. In a TOD, this maximum need never be violated, so it's a reasonable worth value to use.

When such an upper bound does not exist, i.e., an agent is willing to achieve his goal at any cost, other techniques can be used (see Section 6 below).

**Definition 8** *Given an encounter in a two-agent SOD $< s, (G_1, G_2) >$, let $w_i$ be the maximum expected cost that agent $i$ is willing to pay in order to achieve his goal $G_i$. $w_i$ will be called the worth of goal $G_i$ to agent $i$. We will denote this enhanced encounter by $< s, (G_1, G_2), (w_1, w_2) > .$*

The definition of Utility can be usefully altered as follows:

**Definition 9** *Given an encounter $< s, (G_1, G_2), (w_1, w_2) >$, if $\delta$ is a deal, i.e., a mixed joint plan satisfying both agents' goals, then $\text{Utility}_i(\delta)$ is defined to be $w_i - \text{Cost}_i(\delta)$.*

The utility for an agent of a deal is the difference between the worth of its goal that is being achieved, and the cost of his role in the agreed-upon joint plan. If an agent achieves his goal alone, his utility is the difference between the worth of the goal and the cost that he pays to achieve the goal. The point is, that an agent might be better off alone but still derive positive utility from a joint plan, when we use worth as the utility baseline. With the new definition of utility, it may be rational to compromise.

**Theorem 2** *If in Theorem 1 we change every occurrence of $c(s \rightarrow G_i)$ to $w_i$, then that theorem is still true.*

**Proof:** Substitute $w_i$ for every occurrence of $c(s \rightarrow G_i)$ in the proof of Theorem 1. □

By introducing the worth concept into the definition of an encounter, we have enlarged the number of encounters that have a non-empty negotiation set. Cooperative behavior is enhanced. Our negotiation mechanism, that makes use of any product maximizing protocol, becomes applicable to more SOD encounters.

### 4.4 Interaction Types

From the discussion above, we have begun to see emerging different kinds of encounters between agents. In TOD meetings, agents really benefit from coordination. In SODs, this isn't necessarily the case. Sometimes agents benefit, but sometimes they are called upon to bear a coordination overhead so that everyone can achieve their goals. In even more extreme situations, agents' goals may simply be in conflict, and it might just be impossible to satisfy all of them at the same time, or the coordination overhead may exceed the willingness of agents to bear the required burden.

We have four possible interactions, from the point of view of an individual agent:

- A *symmetric cooperative* situation is one in which there exists a deal in the negotiation set that is preferred by both agents over achieving their goals alone. Here, both agents welcome the existence of the other agent.

- A *symmetric compromise* situation is one where there are individual rational deals for both agents. However, both agents would prefer to be alone in the world, and to accomplish their goals alone. Since each agent is forced to cope with the presence of the other, he would prefer to agree on a reasonable deal. All of the deals in NS are better for both agents than leaving the world in its initial state $s$.

- A *non-symmetric cooperative/compromise* situation is one in which one agent views the interaction as cooperative (he welcomes the existence of the other agent), while the second views the interaction as compromise (he would prefer to be alone in the world).

- A *conflict* situation is one in which the negotiation set is empty—no individual rational deals exist.

In a general SOD, all four types of interaction can arise. In a TOD, only the symmetric cooperative situation ever exists.

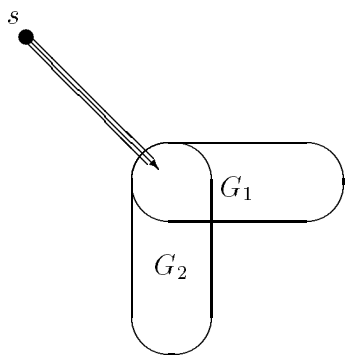

Figure 7: The Symmetric Cooperative Situation

Each of these situations can be visualized informally using diagrams. The symmetric cooperative situation can be seen in Figure 7, the symmetric compromise situation in Figure 8, the non-symmetric cooperative/compromise situation in Figure 9, and the conflict situation in Figure 10. A point on the plane represents a state of the world. Each oval represents a collection of world states that satisfies an agent's goal. $s$ is the initial state of the world. The triple lines emanating from $s$ represent a joint plan that moves the world to a final state. Each of the agents will share in the carrying out of that joint plan. The overlap between ovals represents final states that satisfy the goals of both agents $A_1$ and $A_2$. Informally, the distance between $s$ and either oval represents the cost associated with a single-agent plan that transforms the world to a state satisfying that agent's goal.

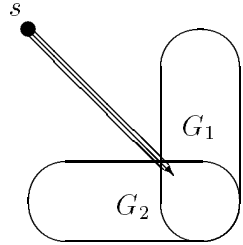

Figure 8: The Symmetric Compromise Situation

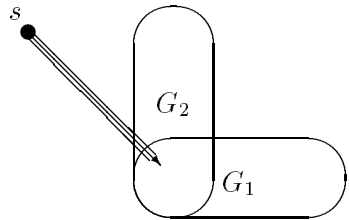

Figure 9: The Non-Symmetric Cooperative/Compromise Situation

Note that in Figure 8, the distance from $s$ to either agent's oval is less than the distance to the overlap between ovals. This represents the situation where it would be easier for each agent to simply satisfy his own goal, were he alone in the world. In Figure 7, each agent actually benefits from the existence of the other, since they will share the work of the joint plan. Note that in Figure 9, one agent benefits from the existence of the other, while the other would prefer to be alone in the world.

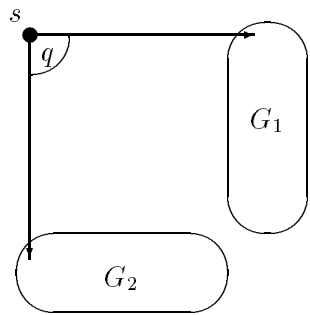

Figure 10: The Conflict Situation

Let's consider some simple examples in the slotted blocks world domain of cooperative, compromise, and conflict situations. In the initial situation depicted in Figure 11, the white block is in slot 1 and the black block is in slot 2. Agent $A_1$ wants the white block alone in slot 2, while agent $A_2$ wants the black block alone in slot 1. Were either of the agents alone in the world, it would cost each of them 4 pickup/putdown operations to achieve their goal. For example, $A_1$ would have to pick up the black block in slot 2 and move it to slot 3, then pick up the white block in slot 1 and move it to slot 2. The two agents together, however, are able to achieve both goals with a *total* cost of 4. They can execute a joint plan where they simultaneously pick up both blocks, and then put them in their appropriate places. Each role in this joint plan costs 2, and each agent derives a utility of 2 from reaching an agreement with the other. This is a cooperative situation. Coordination results in actual benefit for both agents.

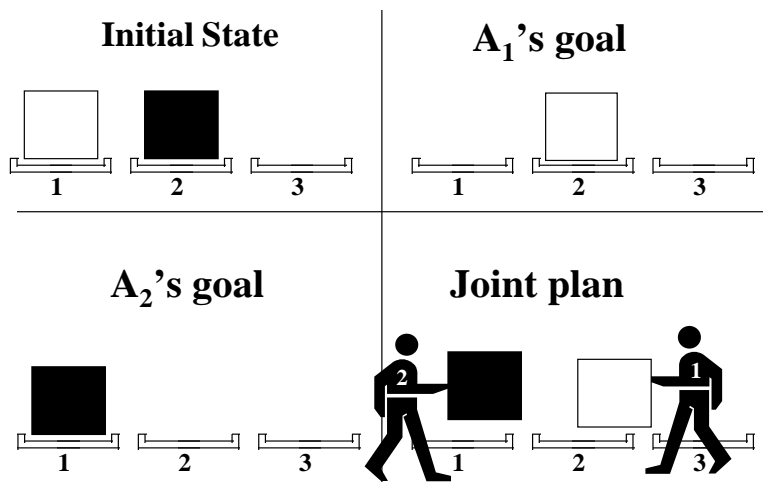

Figure 11: Cooperative Situation

Now let's consider a more complicated, compromise, situation. In the initial state shown in Figure 12, there is a white block in slot 1, a black block in slot 2, and two gray blocks in slot 3. Agent $A_1$'s goal is to have the white block somewhere at slot 2, but not on the table. Similarly, agent $A_2$'s goal is to have the black block somewhere at slot 1, but not on the table. Alone in the world, agent $A_1$ would only have to do one pickup and one putdown operation, just moving the white block onto the black block in slot 2. In the same way, agent $A_2$ alone in the world can achieve his goal with two operations. But since each is (in his stand-alone plan) using the other's block as a base, the achievement of a state that satisfies both agents' goals requires additional blocks and operations.

The best plan for achieving both agents' goals requires moving one gray block from slot 3 to slot 1 and the other gray block to slot 2 so that they act as bases for the white and black blocks. Each block needs to be picked up and put down at least once; the best plan has each block moving only once at a total cost of 8. Obviously, one or both agents will need to do extra work (greater than in the stand-alone situation) to bring about this mutually satisfying state.

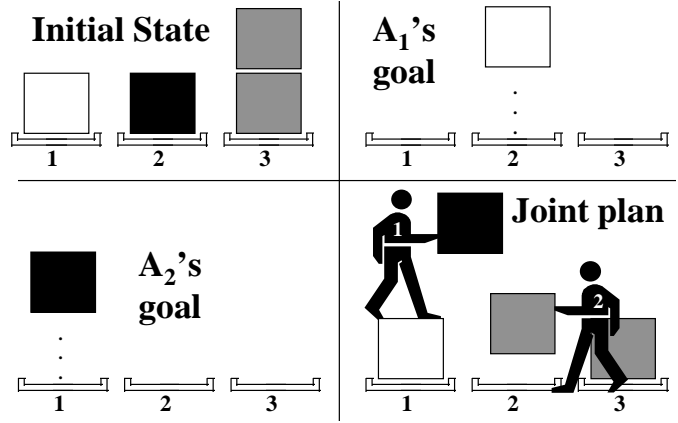

Figure 12: Compromise Situation

The best plan has two roles, one requiring 6 operations and one requiring 2 operations. One agent will lift the black (or white) block, while the other agent rearranges all the other blocks appropriately. The first agent will then put down the black (or white) block, completing the plan. If both agents' worths satisfy the min and sum conditions (meaning, here, that the sum of the worths is greater than or equal to 8, and each worth is greater than or equal to 2), then they can reach an agreement that gives them both positive utility (using worth as the new baseline for evaluating utility).

For example, let's say that agent $A_1$ assigned a worth of 3 to achieving his goal, while agent $A_2$ assigned a worth of 6 to achieving his goal. Since one role in the best joint plan costs 2 while the other costs 6, there is one unit of utility to be shared between the agents. Any mechanism that maximizes the product of their utilities will split this one unit equally between the agents. How is this done in our case? There is one deal in the negotiation set that gives both agents the same expected utility of $\frac{1}{2}$, namely the mixed joint plan that has agent $A_1$ doing the cost-2 role with probability $\frac{7}{8}$, and the cost-6 role with probability $\frac{1}{8}$. Agent $A_2$ of course assumes the complementary role. Agent $A_1$'s expected utility is $3 - 2(\frac{7}{8}) - 6(\frac{1}{8}) = \frac{1}{2}$, which is equal to agent $A_2$'s expected utility of $6 - 2(\frac{1}{8}) - 6(\frac{7}{8}) = \frac{1}{2}$. This division of utility maximizes the product of expected utility among the agents.

There is an interesting phenomenon to note in this deal. Both agents are apparently in a symmetric situation, apart from their internal attitude towards achieving their goals (i.e., how much they are willing to pay). But as can be seen above, the more you are *willing* to pay, the more you will have to pay. The agent that has a worth of 3 ends up having less expected work than the agent with a worth of 6. This gives agents an incentive to misrepresent their true worth values, pretending that the worth values are smaller than they really are, so that the agents' positions within the negotiation will be strengthened. An agent can feign indifference, claim that it really doesn't care all that much about achieving its goal, and come out better in the negotiation (with lower expected cost). We will examine this question in greater detail below in Section 8.

Our final example here is of a conflict situation, shown in Figure 13. Again, the white block is in slot 1 and the black block is in slot 2—the same initial state that we had above in the cooperative and compromise examples. In the cooperative example, the agents wanted the blocks moved to another, empty slot. In the compromise example, the agents wanted to blocks moved to a specific non-empty slot. Here, in the conflict example, the agents want the blocks moved onto a specific other block in a specific slot. Agent $A_1$ wants the white block on top of the black block in slot 2; agent $A_2$ wants the black block on top of the white block in slot 1. Here, there is a real contradiction between the two agents' goals. There exists no world state that satisfies them both. In the next section, we will discuss what kinds of coordination mechanisms can be used in a conflict situation.

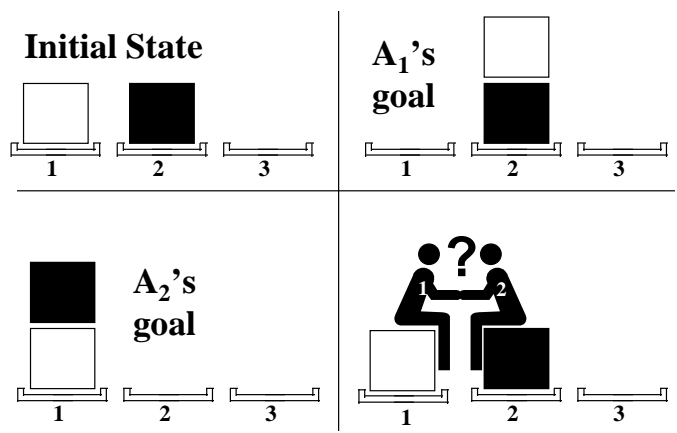

Figure 13: Conflict Situation

When the negotiation set is not empty, we can distinguish between compromise and cooperative situations for a particular agent $i$ using the following algorithm:

1. If $w_i \leq c(s \rightarrow G_i)$, then agent $i$ is in a cooperative situation.

2. If $w_i > c(s \rightarrow G_i)$, then agent $i$ might be in a cooperative or a compromise situation. The way to distinguish between them is as follows:

   (a) Set $w_i^* = c(s \rightarrow G_i)$, and leave the other agent's worths unchanged.

   (b) If the resulting NS* is empty, then agent $i$ is in a compromise situation.

   (c) Otherwise, agent $i$ is in a cooperative situation.

## 5. Conflict Resolution and Cooperation

We have seen that in both cooperative and compromise encounters there exist deals that are individual rational for both agents. Agents will negotiate over which of these deals should be reached, if there is more than one. What, however, can be done when the agents are in a conflict situation, i.e., there are no individual rational deals? Here, the agents have a true conflict that needs to be resolved, and are not merely choosing among mutually acceptable outcomes.

## 5.1 Conflict Resolution

A simple approach to conflict resolution would be for the agents to flip a coin to decide who is going to achieve his goal and who is going to be disappointed. See Figure 14. In this case they will negotiate on the probabilities (weightings) of the coin toss. If they run into a conflict during the negotiation (fail to agree on the coin toss weighting), the world will stay in its initial state $s$.[3]

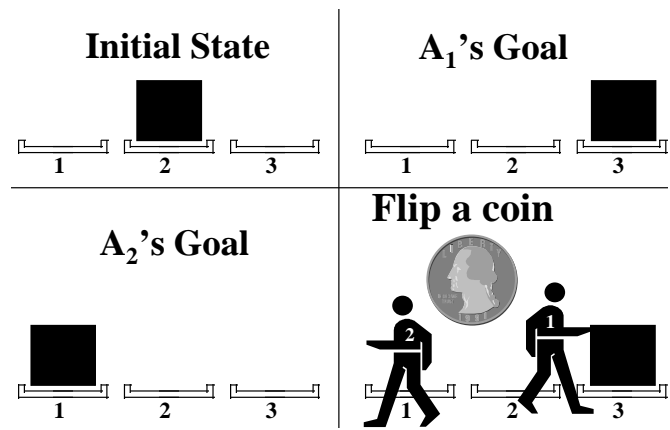

Figure 14: Conflict Resolution

This deal can be visualized graphically in Figure 15. Single lines represent one agent plans.

In conflict situations the agents can use some utility product maximizing protocol to decide on the weighting of the coin. However, it turns out that in that case the probability of $\frac{1}{2}$ always results in the maximum product of the two agents' utilities. If the agents are to maximize their utility product, they will always agree on a symmetric coin. The only exception is when the initial state already satisfies one agent's goal. Then, that agent will simply cause the negotiation to fail, rather than risk moving away from his goal-satisfying state. Nevertheless, even here, the product maximizing deal would have the agents flip a symmetric coin.

Why is a symmetric coin going to maximize the product of agent utilities? Some simple mathematics shows the reason. Assume that agent $A_1$ has a worth of $w_1$, and the cost of achieving his goal alone is $c_1$. When $A_1$ wins the coin toss, he will have a utility of $w_1 - c_1$. His utility from a deal with coin weighting $p$ will be $p(w_1 - c_1)$. His opponent's utility from this deal will be $(1 - p)(w_2 - c_2)$. The product of the two agents' utilities will be $(p - p^2)(w_1 - c_1)(w_2 - c_2)$. This function of $p$ is maximized when $p$ equals $\frac{1}{2}$ for any values of $w_1, w_2, c_1$, and $c_2$ (simply take the derivative of the function and set it equal to zero).

---

3. There is a special case where the initial state $s$ already satisfies one of the agent's goals, let's say agent 1 ($s$ cannot satisfy both goals since then we would not have a conflict situation). In this case, the only agreement that can be reached is to leave the world in state $s$. Agent 1 will not agree to any other deal and will cause the negotiation to fail.

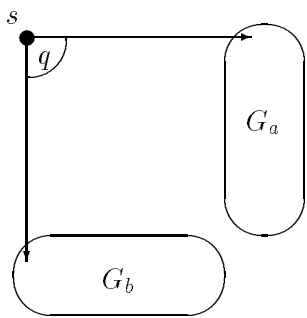

Figure 15: The Conflict Situation

This may seem like a "fair" solution, but it is certainly not an efficient one. While maximizing the product of the agents' utilities, it does not maximize their sum. The sum of utilities will be maximized by simply having the agent with a larger $w_i - c_i$ achieve his goal. This, on the other hand, is certainly not a fair solution.

We might be able to be both fair and efficient if agents are able to transfer utility to one another. In that case, one agent could achieve its goal but share part of that utility with the other agent. The negotiation would then center on how much of the utility should be transferred! Any product maximizing mechanism used to resolve *this* question will transfer half of the gained utility to the other agent, because it is a constant sum game, and dividing the utility equally maximizes the utility product.

The entire subject of explicit utility transfer through side payments is a complicated one that has been treated at length in the game theory community. It is not our intention to examine these questions in this paper. Even if utility is not explicitly transferable, agents can make commitments to perform future actions, and in effect transfer utility through these promises. Again, there are many complicated issues involved in assessing the value of promises, when they should be believed, discount factors, and limits on the amount of promising and debts that an agent can accrue. If agents can accumulate debt indefinitely, it will be possible for them to always pay off previous commitments by making additional commitments to others. Here, too, we are leaving these issues aside, returning to our assumptions that each interaction stands on its own, and no explicit side payments are possible.

## 5.2 Cooperation in Conflict Resolution

In both cooperative and compromise situations, agents were implicitly able to transfer utility in a single encounter by doing more actions in the joint plan. The agent that does more work in the joint plan relieves the other agent, increasing the latter's utility. This can be thought of as a kind of utility transfer. Here, we will see a similar kind of implicit utility transfer that is possible even in conflict situations.

The agents may find that, instead of simply flipping a coin in a conflict situation, it is better for them to cooperatively reach a new world state (not satisfying either of their goals) and *then* to flip the coin to decide whose goal will ultimately be satisfied.

Consider the following example. One agent wants the block currently in slot 2 to be in slot 3; the other agent wants it to be in slot 1. In addition, both agents share the goal of swapping the two blocks currently in slot 4 (i.e., reverse the stack's order). See Figure 16. Assume that $W_1 = W_2 = 12$. The cost for an agent of achieving his goal alone is 10. If the agents decide to flip a coin in the initial state, they will agree on a weighting of $\frac{1}{2}$, which brings them a utility of 1 (i.e., $\frac{1}{2}(12 - 10)$). If, on the other hand, they decide to do the swap cooperatively (at cost of 2 each), and *then* flip a coin, they will still agree on a weighting of $\frac{1}{2}$, which brings them an overall utility of 4 (i.e., $\frac{1}{2}(12 - 2 - 2)$).

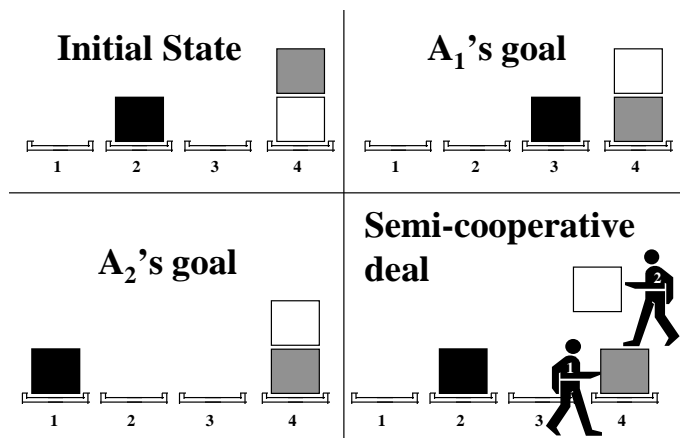

Figure 16: Cooperation up to a Certain Point

The fact that agents, even in a conflict situation, can get more utility by first cooperatively working together, and only then flipping a coin, can be exploited by defining a new kind of deal, called a *Semi-Cooperative Deal*. We want agents to be able to negotiate over and agree on a deal that allows them this mixed cooperative/conflict resolution interaction. Changing the deal type is enough to make this possible. It ends up increasing the expected utility that agents can derive from an encounter.

**Definition 10** *A* Semi-Cooperative Deal *is a tuple* $(t, J, q)$ *where* $t$ *is a world state, $J$ is a mixed joint plan that moves the world from the initial state $s$ to intermediate state $t$, and $0 \leq q \leq 1 \in \mathbb{R}$ is the weighting of the coin toss—the probability that agent $A_1$ will achieve his goal.*

The semantics of this kind of deal is that the two agents will perform the mixed joint plan $J$, and will bring the world to intermediate state $t$. Then, in state $t$, they will flip a coin with weighting $q$ to decide who continues the plan towards their own goal. This allows the agents to handle conflict between their goals, while still cooperating up to a certain point.

The utility of a semi-cooperative deal for an agent can be defined as follows. If he loses the coin toss in intermediate state $t$, he simply has a negative expected utility equal to the expected cost of his role in the joint plan that reached state $t$. If he wins the coin toss in intermediate state $t$, then his expected utility is the difference between the worth of his goal and the costs of his role in the joint plan that reached $t$ as well as the stand-alone cost of moving from $t$ to his goal state. This can be written formally as follows:

**Definition 11**

$$\text{Utility}_i(t, J, q) = q_i(w_i - c(J)_i - c(t \rightarrow G_i)_i) - (1 - q_i)c(J)_i$$
$$= q_i(w_i - c(t \rightarrow G_i)_i) - c(J)_i$$

This assumes, of course, that the agents' goals are in conflict—the state that satisfies one agent will be of no worth to the other.

The Semi-Cooperative Deal can be visualized graphically in Figure 17. This figure is similar in spirit to the figures presented above in Section 4.4, which represented cooperative, compromise, and conflict encounters. Again, a triple line represents a joint plan while a single line represents a one-agent plan.

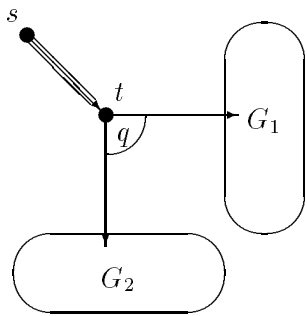

Figure 17: Semi-Cooperative Deal

## 5.3 Semi-Cooperative Deals in Non-Conflict Situations

In cooperative and compromise situations, the agents negotiate on deals that are mixed joint plans, $J : p$ (cooperative). In a conflict situation, the agents negotiate on deals of the form $(t, J, q)$ (semi-cooperative deals).

Even though semi-cooperative deals were intended to be used in conflict situations, they can also be used in cooperative and compromise situations (with a minor generalization in the definition of utility, discussed below). The question is, what kinds of agreements will agents in a non-conflict situation reach, if they are negotiating over semi-cooperative deals? Will they do better than using standard cooperative (mixed joint plan) deals? Or will they do worse?

A *cooperative* deal which is a mixed joint plan $J : p$ can also be represented as $(J(s), J : p, 0)$ where $J(s)$ is the final world state resulting from the joint plan $J$ when the initial state is $s$. In other words, mixed deals are a proper subset of semi-cooperative deals, and any mixed deal can be represented as a semi-cooperative deal of a special form. The intermediate state $t$ is taken to be the final state of the agents' cooperative joint plan. Since that final state satisfies both agents' goals, the result of the coin flip is irrelevant—neither of the agents wants to change the world state anyway.

Therefore, by having agents in a non-conflict situation negotiate over semi-cooperative deals, we are only enlarging their space of agreements. Any deal that can be reached when negotiating over the subset (i.e., mixed joint plans) can also be reached when negotiating over the larger set (i.e., semi-cooperative plans). So the agents in a non-conflict situation will certainly do no worse, when using semi-cooperative deals. But will they do better?

There are two potential ways in which agents could do better. The first would be if agents find a cheaper way to achieve both goals. It turns out that this is impossible—semi-cooperative deals will not uncover a more efficient way of achieving both agents' goals. However, there is a more surprising way in which agents can benefit from semi-cooperative deals. Agents can benefit by not always achieving their goals. When using semi-cooperative deals, they can give up guaranteed goal satisfaction, and gain expected utility.

To see what we mean, consider the following example in the Slotted Blocks World. The initial situation in Figure 18 consists of 5 duplications of the example from Figure 5, in slots 1 to 15. In addition, two slots (16 and 17) each contain a stack of 2 blocks. Agent $A_1$'s goal is "White blocks are in slots $2, 5, 8, 11$ and 14 but not on the table; the blocks in slots 16 are swapped, and the blocks in slot 17 are swapped (i.e., each tower is reversed)." Agent $A_2$'s goal is "Black blocks are in slots $1, 4, 7, 10$ and 13 but not on the table; the blocks in slot 16 are swapped, and the blocks in slot 17 are swapped."

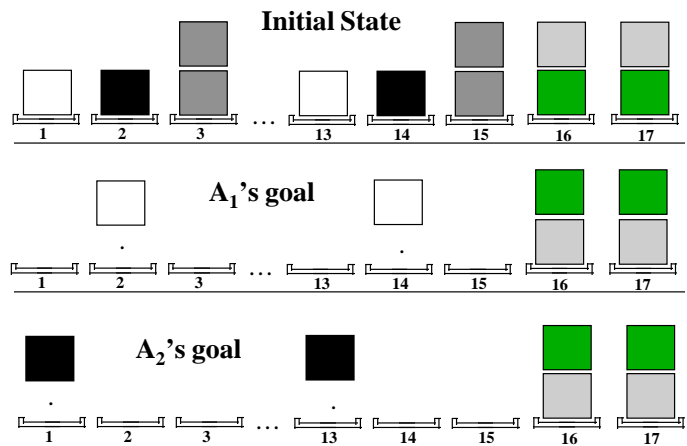

Figure 18: Semi-Cooperative Agreement in a Cooperative Situation

The stand-alone cost of both agents is: $c(s \rightarrow G_i) = 26 = (5 \times 2) + (2 \times 8)$. Let's assume that $w_i = 26$ is also the worth of each goal. The minimal cost joint plan that achieves both agents' goals has 7 parts:

- Cooperative swap in slot 17 in which each agent does one pickup and one putdown;

- The same swap in slot 16;

- Five duplications of the joint plan from Example 5. Each of those joint plans has a role that costs 6 and a role that costs 2.

Thus the average cost of each agent's role in the joint plan is 24, namely $(2 \times 2) + (5 \times \frac{1}{2}(6 + 2))$. Since the stand-alone cost is 26, this situation is cooperative—each agent welcomes the existence of the other. For both agents, the expected utility of the joint plan is 2 (i.e., $26 - 24$). This cooperative deal achieves both agents' goals.

Can we find a semi-cooperative deal that is better? What if the agents only cooperated on swapping the blocks in slots 16 and 17, and then tossed a coin to see who gets to fulfill his own goal (leaving the other's goal unsatisfied)? This semi-cooperative deal actually turns out to be better for both agents.

Let the intermediate state $t$ be the state where the blocks in slots 16 and 17 are swapped, and the other slots are unchanged. Each agent invests 4 operations as his part of the two swaps. He then has a chance of $\frac{1}{2}$ of continuing on his own to achieve his goal, and a chance of $\frac{1}{2}$ of losing the coin toss and having wasted his initial investment. If he wins the coin toss, he will have an additional 10 operations ($5 \times 2$), but achieve his goal of worth 26. Overall utility will be $26 - 10 - 4$, i.e., 12. If he loses the coin toss, he has just wasted his initial investment of 4, so his utility is $-4$. The expected utility is the average of these two cases, i.e., 4. This is better than the utility of 2 the agents got using the cooperative deal!

In other words, in this case, the agents would prefer *not* to guarantee themselves their goal, and take a gamble with a semi-cooperative deal. Their expected utility doubles, if they are willing to take a risk. So even in this cooperative situation, the agents benefit from negotiating over semi-cooperative deals.

Now, it turns out that this is a borderline situation, brought about because $w_i$ is low. As long as $w_i$ is high enough, any semi-cooperative deal that agents agree on in a cooperative situation will be equivalent to a cooperative deal. If achieving your goal isn't worth too much to you (your profit margin is small), you might be willing to forgo guaranteed achievement in exchange for a higher expected utility.

So semi-cooperative deals, used in a non-conflict situation, will sometimes result in better agreements (when forgoing guaranteed goal achievement is beneficial), and will never result in worse agreements. Clearly, it is worthwhile for agents to negotiate over semi-cooperative deals, regardless of whether they are in cooperative, compromise, or conflict situations.

## 5.4 Unified Negotiation Protocols (UNP)

In this section, we will make the necessary generalizations so that agents can use semi-cooperative deals in all types of encounters. We will call all product maximizing mechanisms based on semi-cooperative deals "Unified Negotiation Protocols (UNP)," since they can be used for conflict resolution, as well as for cooperative agreements.[4]

As mentioned above, we will need to generalize the previous definition of the utility of a semi-cooperative deal, to enable UNPs. Before, we assumed (since we had a conflict situation) that the final state would be of no benefit to the agent that lost the coin toss.

---

4. An earlier version of this subsection and the next two appeared in (Zlotkin & Rosenschein, 1991a). The current treatment incorporates new material on multi-plan deals, recasts the protocols in the context of domain theory, and alters the notation to correspond to the more general domain framework.

Now, even though a semi-cooperative deal is being used, the final state might still satisfy both agents' goals, and not just the goal of the agent that wins the coin toss.

If $(t, J, q)$ is a semi-cooperative deal, then we'll define $f_i$ to be the final state of the world when agent $i$ wins the coin toss in state $t$. $f_i = (t \rightarrow G_i)(t) \in G_i$. The worth for agent $i$ of any state $r$, which we will write as $W_i(r)$, will be his goal worth $w_i$ if $r$ is a goal state, and 0 otherwise. Now, we can revise our definition of utility for semi-cooperative deals:

**Definition 12** Utility$_i(t, J, q) = q_i(w_i - c(t \rightarrow G_i)_i) + (1 - q_i)w_i(f_j) - c(J)_i$

The utility of a semi-cooperative deal $(t, J, q)$ for an agent is now defined to be the expected worth of the final state minus the expected cost. The worth of the expected final state, of course, depends on the weighting of the coin, and whether both possible final states (or only one) are goal states for the agent. Similarly, the expected cost depends on the weighting of the coin (whether the agent only participates in the first, joint, plan, or also continues with the second, lone, plan).

The definition of utility given above is completely consistent with the earlier definition of the utility of cooperative deals, and can be viewed as a generalization of that earlier definition. In other words, if a cooperative deal (a mixed joint plan) is mapped into a semi-cooperative deal $(t, J, q)$ using the transformation discussed above, then the definition of utility for a mixed joint plan (Definition 9) and this definition of utility (Definition 12) for a semi-cooperative deal yield the same number.

A sufficient condition for the negotiation set to be non-empty over semi-cooperative deals is that agents' worths be high enough, so that each agent would be able to achieve its own goal alone:

**Theorem 3** *If for each agent $i$ the worth of its goal is greater than or equal to his stand-alone cost (i.e., $\forall i \ w_i \geq c(s \rightarrow G_i)$), then the negotiation set over semi-cooperative deals is not empty.*

**Proof:** To show that NS $\neq \emptyset$, it is sufficient to show that there is an individual rational semi-cooperative deal. The existence of pareto-optimal deals among the individual rational deals is due to the compactness of the deal space (since there is only a finite number of agent operations, and the worth of agent goals is bounded). $(s, \Lambda, q)$, where $\Lambda$ is the empty joint deal, is individual rational for any $q$. $\quad\Box$

The above condition is sufficient, but is not necessary, for the negotiation set to be non-empty. For example, consider again the situation given in Figure 16, but with the agents' worths being equal to 8 (instead of 12). Neither agent can achieve its goal alone, and thus the conditions of the theorem above are not satisfied. However, there is a perfectly good semi-cooperative deal that gives both agents positive utility—they perform a joint plan that swaps the blocks in slot 4, then flip a coin to see whether the block in slot 2 goes to slot 1 or 3. This mixed deal gives each agent an expected utility of 1. So the negotiation set is not empty.

It turns out that if there is a semi-cooperative deal in the negotiation set, and one of the agents, winning the coin toss, will bring the world into a state that satisfies *both* agents' goals, then there exists another deal in the negotiation set with the same utility for both agents in which the intermediate state already satisfies both agents' goals.

**Theorem 4** *For a semi-cooperative deal $(t, J, q) \in \text{NS}$, if there exists an $i$ such that $f_i \in G_1 \cap G_2$, then this semi-cooperative deal is equivalent to some cooperative deal.*

**Proof:** There are two cases: *both* final states, or *only one* final state, is in $G_1 \cap G_2$.

- If $f_1, f_2 \in G_1 \cap G_2$, then we can view the last step as performing a mixed joint plan that moves the world from state $t$ to a state in $G_1 \cap G_2$.

$$c(t \rightarrow G_1) = c(t \rightarrow G_1 \cap G_2) = c(t \rightarrow G_2),$$

  because if $X \subset Y$ and $(t \rightarrow Y)(t) \in X$, then $c(t \rightarrow X) = c(t \rightarrow Y)$. $f_1, f_2$ are not necessarily the same state, but this deal is equivalent to a deal where $f_1 = f_2$. We can look at the concatenation of the two mixed joint plans (the first being $J$ from $s$ to $t$, the second $t \rightarrow G_1 \cap G_2$), as a mixed joint plan $P$ from $s$ to $G_1 \cap G_2$. $P$ is a cooperative deal that is equivalent to $(t, J, q)$, because

$$\begin{aligned}
\text{Utility}_i(t, J, q) &= q_i(w_i - c(t \rightarrow G_i)) + (1 - q_i)(w_i) - c(J)_i \\
&= w_i - (q_i c(t \rightarrow G_i) + c(J)_i) \\
&= w_i - c(P)_i \\
&= \text{Utility}_i(P).
\end{aligned}$$

- If $f_1 \in G_1 \cap G_2$ and $f_2 \notin G_1 \cap G_2$, then agent 2 would prefer to lose the coin toss at state $t$ and let agent 1 achieve the goal *for* him without his spending any more. The deal $(t, J, 1)$ is better for 1 and can only be better for 2 as well, so it dominates $(t, J, q)$; but $(t, J, q) \in \text{NS}$, so they are equivalent. $(t, J, 1)$ is equivalent to the mixed joint plan $P$ in which the agents perform the joint plan $J$ until $t$, and then agent 1 performs the one-agent plan $t \rightarrow G_1 \cap G_2$. $P$ is a cooperative deal.

$\square$

In other words, if there exists a semi-cooperative deal in the negotiation set that *sometimes* satisfies both agents' goals (depending on who wins the coin toss), then there also exists another semi-cooperative deal in the negotiation set that *always* satisfies both agents' goals (equivalent to a cooperative deal). Even though semi-cooperative deals constitute a superset of cooperative deals, no extra utility is derived from using semi-cooperative deals if the agreement preserves mutual satisfaction of both agents (i.e., if it's equivalent to a cooperative deal).

In a cooperative situation, agents cannot extract more utility from a semi-cooperative deal, unless they are willing to agree on a deal that will *never* satisfy both agents' goals. The example above (Section 5.3) is the prototype of that situation. Agents increase their utility by using a semi-cooperative deal in a cooperative situation. They do this by forgoing guaranteed mutual satisfaction. The above theorem implies that this is the only way they can increase their utility with semi-cooperative deals.

## 5.5 Multi-Plan Deals

In semi-cooperative deals, we assume that the agents cooperate, flip a coin, then the winner proceeds alone to achieve his goal. This arrangement only requires that the agents engage in "pre-flip cooperation." What if the agents were willing (or required) to also engage in "post-flip cooperation"? Then, an entirely new dimension of agreements would be opened up. In this section, we consider a kind of deal that exploits cooperation after the coin toss.

To illustrate the potential of this new kind of deal, consider the following encounter, shown in Figure 19.

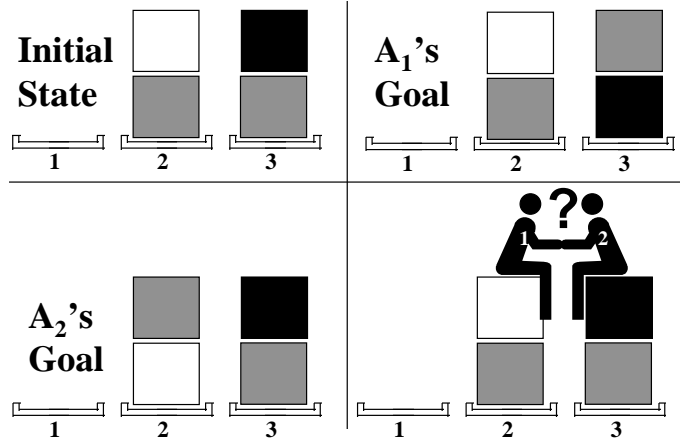

Figure 19: Post-Flip Cooperation can be Helpful

The initial state $s$ of the world can be seen in Figure 19. $A_1$'s goal is to swap the position of the blocks in slot 3, but to leave the blocks in slot 2 in their initial position. $A_2$'s goal is to swap the position of the blocks in slot 2, but to leave the blocks in slot 3 in their initial position.

To achieve his goal alone, each agent needs to do at least 8 pickup or putdown operations. Apparently, there is very little room for cooperation. Not only is there no final state that satisfies both agents' goals, there is no intermediate state (other than the initial state) to which the agents can cooperatively bring the world, before tossing a coin (as in a semi-cooperative deal).

Negotiating over semi-cooperative deals, agents will agree to flip a coin in the initial state, and whoever wins the coin toss will by himself bring the world into his goal state (at a cost of 8). Assume that the worth of each agent's goal is 10. Then negotiating over semi-cooperative deals brings each agent an expected utility of 1. This is a compromise situation (alone in the world, each agent would have utility of 2).

What if the agents could reach the following agreement (as shown in Figure 20): they flip a coin in the initial state. Whoever wins the toss gets his goal satisfied. However, no matter who wins, the agents commit themselves to work together in a joint plan to achieve the chosen goal.

Doing either swap jointly costs a total of 4 for the two agents (2 each). The agent that wins the coin toss gets a utility of $10 - 2$ (his goal is satisfied and he expends 2 in the joint plan). The agent that loses gets a utility of $-2$ (he just expends 2 in the joint plan that achieves his opponent's goal). If each agent has an equal chance of winning the coin toss, his expected utility will be 3. This is better than the semi-cooperative deal that gave

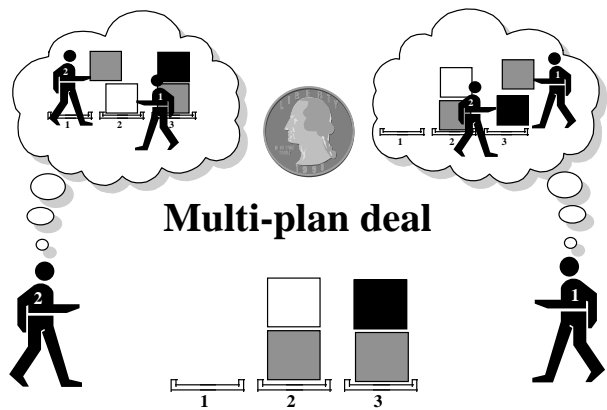

**Multi-plan deal**

Figure 20: Multi-Plan Deal

the agents each a utility of 1. It's even better than the stand-alone utility of 2 that the agents could get if they were alone! Suddenly, the situation has become cooperative. The agents welcome each other's existence, even though their goals have nothing in common. There is no goal state that satisfies both agents; there are no subgoals that the agents have in common; there are no positive interactions between the agents' stand-alone plans. The goals are completely decoupled, and yet the situation is cooperative.

The agreement above, of course, requires "post-flip cooperation." With semi-cooperative deals, the "pre-flip cooperation" contributed potentially to either agents' benefit—either agent might win the coin toss and exploit the early work. But with this new deal type, even the agent who loses the coin toss will be required to expend effort, knowing that it is just for the benefit of the other agent.

If agents will commit themselves to post-flip cooperation, then this new deal type is possible. Agents could then negotiate over deals that are pairs of mixed joint plans. We will call these new deals *multi-plan deals*. By committing to post-flip cooperation, agents enlarge the space of agreements, and this potentially improves their expected utility.

**Definition 13**

- *A Multi-Plan Deal is $(\delta_1, \delta_2, q)$, where each $\delta_i$ is a mixed joint plan that moves the world to a state that satisfies $i$'s goal. $0 \le q \le 1 \in \mathbb{R}$ is the probability that the agents will perform $\delta_1$ (they will perform $\delta_2$ with probability $1 - q$).*

- *Assuming $j$ is $i$'s opponent, we have $\text{Utility}_i(\delta_1, \delta_2, q) = q(w_i - \text{Cost}_i(\delta_i)) - (1 - q)\text{Cost}_i(\delta_j)$.*

So a multi-plan deal has agents agreeing on two joint plans, and deciding which to execute by tossing a coin. This deal can be visualized informally in Figure 21, as in Section 4.4 above. A triple line represents a joint plan, carried out by multiple agents.

Note that here the symmetric abilities assumption from Section 2.4 may not be essential (i.e., with the multi-plan deal type agents may not need to be able to perform the same plans at equivalent cost). The two mixed joint plans that comprise a multi-plan deal might be pure (i.e., $p$ can be 0 or 1) without overly restricting the agents' ability to divide the utility accurately, since the agents have the additional $q$ probability that they can adjust.

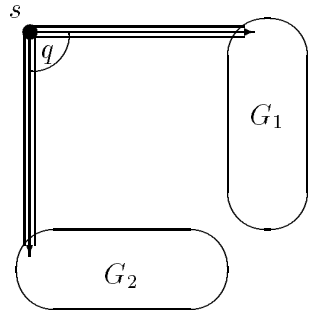

Figure 21: Multi-Plan Deal

Just as semi-cooperative deals can be used in cooperative situations, multi-plan deals can also be used in cooperative situations (since, as we will see below, they are a generalization of semi-cooperative deals). All that is needed is to enhance the definition of multi-plan deal utility appropriately, as was done for semi-cooperative deals (Definition 12).

Consider the following example, which shows the increased utility available for the agents to share when they negotiate over multi-plan deals instead of over mixed joint plans.

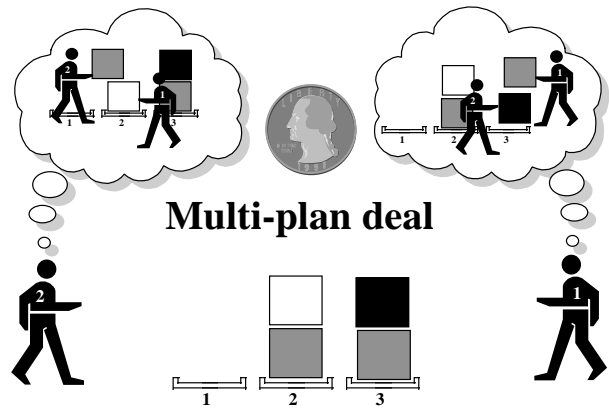

Figure 22: Relationship of the Multi-Plan Deal Type to Mixed Joint Plans

The initial state $s$ of the world can be seen in Figure 22. $A_1$'s goal is to swap the position of the blocks in slot 3, but to leave the blocks in slot 2 in their initial position (there is only one state that satisfies this goal; call it $f_1$). $A_2$'s goal is to swap the position of the blocks in slot 2, but to leave the blocks in slot 3 in their initial position ($f_2$).

To achieve his goal alone, each agent needs to do at least 8 pickup or putdown operations (each with a cost of 1). Assume that $A_1$'s worth function assigns 10 to $f_1$ and 0 to all other states, and that $A_2$'s worth function assigns 10 to $f_2$ and 0 to all other states. In this case, the negotiation set includes the deals $(s \to f_1, \Lambda)\!:\!1$ and $(\Lambda, s \to f_2)\!:\!0$. Using the protocol mentioned above, the agents will break the symmetry of this situation by flipping a coin. The utility of each agent from this deal is $1 = \frac{1}{2}(10 - 8)$.

Negotiation over the multi-plan deal type will cause the agents to agree on $(\delta_1, \delta_2)\!:\!\frac{1}{2}$, where $\delta_i$ is the mixed joint plan in which both agents cooperatively achieve $i$'s goal. The best joint plan for doing the swap in one of the slots costs 2 pickup/putdown operations for each agent. The utility for each agent from this deal is $3 = (\frac{1}{2}(10 - 2) + \frac{1}{2}(-2))$. By negotiating using the multi-plan deal type instead of mixed joint plans, there is more utility for the agents to share, 6 instead of 2.

### 5.6 The Hierarchy of Deal Types — Summary

There exists an ordering relationship among the various kinds of deals between agents that we have considered; we call this relationship the "deal hierarchy." At the bottom of the hierarchy are *pure deals* and *mixed deals*. These first two types of deals in the hierarchy can be used only in cooperative situations. For negotiation in general non-cooperative domains, additional types of deals were needed.

Next in the hierarchy come *semi-cooperative deals*. As we have shown, semi-cooperative deals are a superset of mixed deals. Even in cooperative situations, there may be some semi-cooperative deals that do not achieve all goals, but which dominate all other mixed joint plans that *do* achieve both agents' goals.

Finally, at the top of the hierarchy, come *multi-plan deals*, which are a superset of semi-cooperative deals. This is the most general deal type in our deal hierarchy. This deal type can also serve as the foundation for a class of Unified Negotiation Protocols.

In summary, our hierarchy looks as follows:

$$\{J\} \subseteq \{J\!:\!p\} \subseteq \{t, \delta, q\} \subseteq \{(\delta_1, \delta_2)\!:\!q\}$$

Pure Deals $\subseteq$ Mixed Deals $\subseteq$ Semi-Cooperative Deals $\subseteq$ Multi-Plan Deals

## 6. Unbounded Worth of a Goal—Tidy Agents

In Section 4.3, we assumed that each agent assigns a finite worth to achieving his goal, which is the upper bound on cost that he is willing to spend to achieve the goal. What if such an upper bound does not exist? There may be situations and domains in which there is no limit to the cost that an agent is willing to pay in order to achieve his goal—he would be willing to pay *any* cost. Similarly, there may be situations when an agent is simply unable, by design, to evaluate the worth of its goal. However, even though the worth is unbounded or unevaluable, the agent is still interested in expending the minimum necessary to achieve its goal. The agent gets more utility when it spends less, and can determine an ordinal ranking over all possible deals, even though it has difficulty assigning cardinal values to the utility derived from those deals.

Nevertheless, we really are interested in having cardinal values that can be used in our negotiation mechanisms. Our whole approach to negotiation is founded on the existence of these inter-agent comparable cardinal utility functions. When worth is unbounded for both agents, we seem to be deprived of the tool on which we have depended.

We would like to identify a different baseline by which to define the concept of utility. Originally, above, we used the baseline of "stand-alone cost," $c(s \rightarrow G_i)$, taking the utility of a deal for an agent to be the difference between the cost of achieving the goal alone and the agent's part of the deal. Then, we used the baseline of "worth" in a similar manner, linearly transforming the utility calculation. Utility of a deal for an agent was then the difference between the maximum cost he was willing to pay and the agent's part of the deal. When worth is unbounded, however, that linear transformation obviously cannot be used.

In other work (Zlotkin & Rosenschein, 1993b), we present an alternative baseline that can satisfy our desire for symmetry, fairness, simplicity, stability, and efficiency. It turns out to constitute the minimum sufficient baseline for agents to reach agreements.

The minimum cost that an agent must offer to bear in a compromise encounter, where neither agent has an upper bound on its worth, is that which leaves the other agent with less cost than the latter's stand-alone cost. In other words, the first agent will offer to "clean up after himself," to carry out a sufficient portion of the joint plan that achieves both goals such that the other agent's remaining part of the joint plan will cost him less than his stand-alone cost. We call an agent who is willing to clean up after himself a *tidy* agent; the formal definition appears elsewhere (Zlotkin & Rosenschein, 1993b). It is shown that in any joint-goal reachable encounter (i.e., there exists a joint plan that achieves both agents' goals), if both agents are tidy, the negotiation set is not empty.

## 7. Negotiation with Incomplete Information

All the mechanisms considered in the sections above can be straightforwardly implemented only if both agents have full information about each other's goals and worths. In many situations, this won't be the case, and in this section we will examine what happens to our negotiating mechanisms in State Oriented Domains when agents don't necessarily have full information about each other.

We consider incomplete information about goals, and incomplete information about worths, as two separate issues. An agent, for example, might have particular information about worth, but not about goals, or vice versa. There are thus four possible cases, where worths are known or not known, combined with goals that are known or not known. In previous sections, we considered the case where both goals and worth were known. In this section we consider two of the other three situations, where neither goals nor worth are known, and where goals are known and worth is not. We do not analyze situations where worth is known but the goals are not.

The general conclusion is that a strategic player can gain benefit by pretending that its worth is lower than it actually is. This can be done directly, by declaring low worth (in certain mechanisms), or by declaring a cheaper goal (in the case where stand-alone cost is taken to be the implicit worth baseline).

In this first section, we consider the space of lies that are available in different types of interactions, and with different types of mechanisms.

There are several frameworks for dealing with incomplete information, such as incremental goal recognition techniques (Allen, Kautz, Pelavin, & Tenenberg, 1991), but the framework we explore here is that of a "−1 negotiation phase" in which agents simultane-ously declare private information before beginning the negotiation (this was also introduced elsewhere (Zlotkin & Rosenschein, 1989, 1993a) for the case of TODs). The negotiation then proceeds as if the revealed information were true. In the TOD case, we have analyzed the strategy that an agent should adopt for playing the extended negotiation game, and in particular, whether the agent can benefit by declaring something other than his true goal. Here, we will take a similar approach, and consider the $-1$-phase game in State Oriented Domains. Will agents benefit by lying about their private information? What kinds of mechanisms can be devised that will give agents a compelling incentive to only tell the truth?[5]

A negotiation mechanism that gives agents a compelling incentive to only tell the truth is called (in game theory) *incentive compatible*. Although we are able to construct an incentive compatible mechanism to be used when worths are unknown, we are unable to construct such a mechanism in State Oriented Domains to be used when the other's goals are unknown.

## 7.1 Worth of a Goal and its Role in Lies

We again assume that agents associate a *worth* with the achievement of a particular goal. Sometimes, this worth is exactly equal to what it would cost the agent to achieve that goal by himself. At other times, the worth of a goal to an agent exceeds the cost of the goal to that agent. The worth of a goal is the baseline for calculating the utility of a deal for an agent; in this section, we will always assume that worth is bounded.

The worth of a goal is intimately connected with what specific deals agents will agree on. First, an agent will not agree on a deal that costs him more than his worth (he would have negative utility from such a deal). Second, since agents will agree on a deal that maximizes the product of their utilities, if an agent has a lower worth, it will ultimately reduce the amount of work in his part of the deal. Thus, one might expect that if agent $A_1$ wants to do less work, he will try to fool agent $A_2$ into thinking that, for any particular goal, $A_1$'s worth is lower than it really is. This strategy, in fact, often turns out to be beneficial, as seen below.

Let's consider the following example from the Slotted Blocks World.

The initial state can be seen at the left in Figure 23. $G_1$ is "The Black block is on a Gray block which is on the table at slot 2" and $G_2$ is "The White block is on a Gray block which is on the table at slot 1".

To achieve his goal alone, each agent has to execute four PickUp and four PutDown operations that cost (in total) 8. The two goals do not contradict each other, because there exists a state in the world that satisfies them both. There also exists a joint plan that moves the world from the initial state to a state that satisfies *both* goals with total cost of 8—one agent lifts the black block, while the other agent rearranges the other blocks suitably (by picking up and putting down each block once), whereupon the black block is put down. The agents will agree to split this joint plan with probability $\frac{1}{2}$, leaving each with an expected utility of 4.

---

[5]. Some of these issues, in everyday human contexts, are explored in (Bok, 1978). Our immediate motivation for discouraging lies among agents is so that our negotiation mechanisms will be efficient.

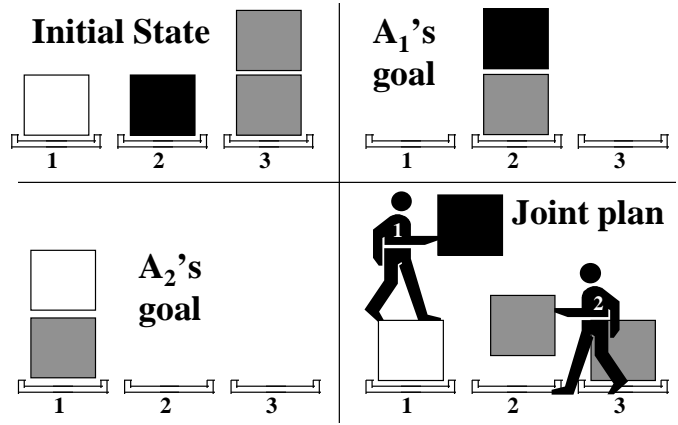

Figure 23: Agents Work Together Equally

## 7.2 Beneficial Lies with Mixed Deals

What if agent $A_1$ lies about his true goal above, claiming that he wants a black block on *any* other block at slot 2? See Figure 24. If agent $A_1$ were alone in the world, he could apparently satisfy this relaxed goal at cost 2. Assuming that agent $A_2$ reveals his true goal, the agents can only agree on one plan: agent $A_1$ will lift a block (either the white or black one), while agent $A_2$ does all the rest of the work. The apparent utility for agent $A_1$ is then 0 (still individual rational), while agent $A_2$ has a utility of 2. In reality, agent $A_1$ has an actual utility of 6. Agent $A_1$'s lie has benefited him.

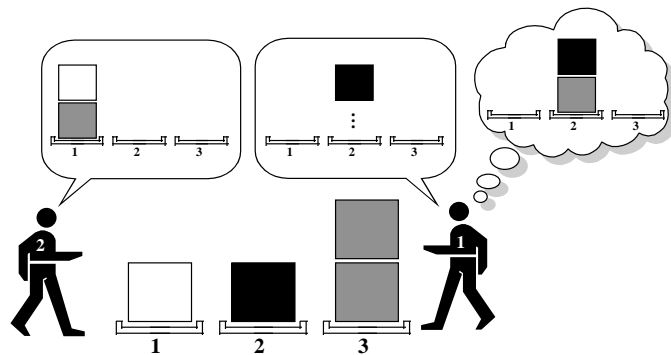

Figure 24: Agent $A_1$ Relaxes his Goal

This works because agent $A_1$ is able to reduce the apparent cost of his carrying out his goal alone (which ultimately causes him to carry less of a burden in the final plan), while not compromising the ultimate achievement of his real goal. The reason his real goal is "accidentally" satisfied is because there is only one state that satisfies agent $A_2$'s real goal and agent $A_1$'s apparent goal, coincidentally the same state that satisfies both of their real goals.

The lie above is not agent $A_1$'s only beneficial lie in this example. What if agent $A_1$ claimed that his goal is "Slot 3 is empty and the Black block is clear"? See Figure 25. Interestingly, this goal is quite different from his real goal. If agent $A_1$ were alone in the world, he could apparently satisfy this variant goal at cost 4. The agents will then be forced again to agree on the deal above: $A_1$ does two operations, with apparent utility of 2, and agent $A_2$ does six operations, with utility of 2. Again, agent $A_1$'s actual utility is 6.

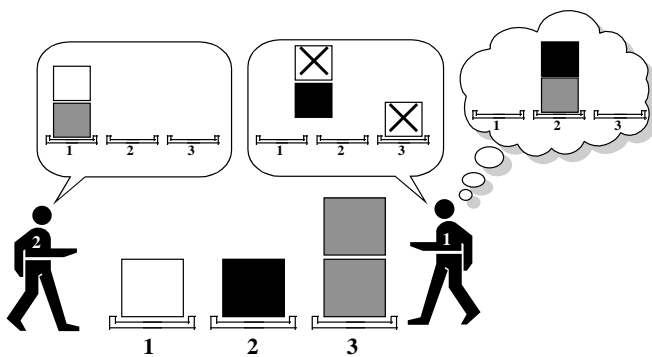

Figure 25: Agent $A_1$ Makes Up an Entirely New Goal

In Task Oriented Domains (Zlotkin & Rosenschein, 1989, 1993a), we also saw something similar to this lying about a goal. There, for example, an agent could hide a task, and lower the apparent cost of its stand-alone plan. Similarly, in the first lie above the agent in the Blocks World relaxed his true goal, and lowered the apparent cost of his stand-alone plan (and thus his worth). The set of states that will satisfy his relaxed goal is then a superset of the set of states satisfying his true goal.

However, there is a major difference between lying in SODs and lying in TODs: in the latter, there can never be any "accidental" achievement of hidden goals. The lying agent will always find it necessary to carry out the hidden goal by himself, and this is the main reason why in subadditive TODs hiding goals is not beneficial. In SODs, a hidden goal might be achieved by one's opponent, who carries out actions that have side effects. Thus, even when you hide your goal, you may fortuitously find your goal satisfied in front of your eyes.

This situation can be visualized informally in Figure 26, as other SOD interactions were in Section 4.4 above. In the figure, agent $A_1$'s expanded apparent goal states are represented by the thicker oval and labeled $G_1'$. Note that the expansion of the goal states is toward the initial state $s$. This is the meaning of lowering one's apparent cost, and is necessary for a beneficial lie.

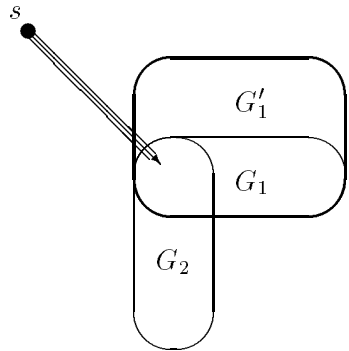

Figure 26: Expanding Apparent Goal States with a Lie

Alternatively, the agent can manufacture a totally different goal for the purposes of reducing his apparent cost, as we saw in Figure 25. Agent $A_1$ did this when he said he wanted slot 3 empty and the Black block clear. Consider Figure 27, where agent $A_1$'s altered apparent goal states are again represented by the thick outline and labeled $G_1'$. Note again, that the expansion of the goal states is toward the initial state $s$.

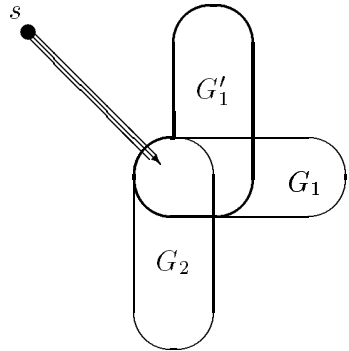

Figure 27: Altering Apparent Goal States with a Lie

The agent then needs to make sure that the intersection of his apparent goal states and his true goal states is not empty. Although this is a necessary precondition for a successful lie, it is of course not a sufficient precondition for a successful lie. Both of the lies in the above example will be useful to agent $A_1$ regardless of the negotiation protocol that is being used: pure deal, mixed deal, semi-cooperative deal, or multi-plan deal.

## 7.3 Beneficial Lies with Semi-Cooperative Deals

It might seem that when agents are in a conflict situation, the potential for beneficial lies is reduced. In fact, beneficial lying *can* exist in conflict situations.

"Conflict" between agents' goals means that there does not exist a mixed joint plan that achieves both goals and is also individual rational. This is either because such a state does not exist, or because the joint plan is too costly to be individual rational. Even when conflict exists between goals, there might be common subgoals, and therefore a beneficial lie may exist.

**Taking Advantage of a Common Subgoal in a Conflict Situation:** Let the initial state of the world be as in Figure 28. One agent wants the block currently in slot 2 to be in slot 1; the other agent wants it to be in slot 3. In addition, both agents share the goal of swapping the two blocks currently in slot 4 (i.e., reverse the stack's order).

The cost for an agent of achieving his goal alone is 10. Negotiating over the true goals using semi-cooperative deals would lead the agents to agree to do the swap cooperatively (at cost of 2 each), and then flip a coin, with a weighting of $\frac{1}{2}$, to decide whose goal will be individually satisfied. This deal brings them an overall expected utility of 2 (i.e., $\frac{1}{2}(10 - 2) - 2$).

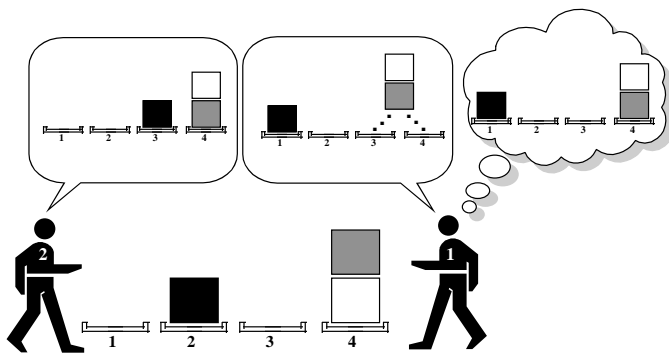

Figure 28: Taking Advantage of a Common Subgoal

What if agent $A_1$ lies and tells agent $A_2$ that his goal is: "The Black block is clear at slot 1 and the White block is on the Gray block"? Agent $A_1$ thus hides the fact that his real goal has the stack of blocks at slot 4, and claims that he does not really care if the stack is at slot 2, 3 or 4. The cost for agent $A_1$ of achieving his apparent goal is 6, because now he can supposedly build the reversed stack at slot 3 with a cost of 4. Assuming that agent $A_2$ reveals his true goal, the agents will still agree to do the swap cooperatively, but now the weighting of the coin will be $\frac{4}{7}$. This deal would give agent $A_1$ an apparent utility of $1\frac{3}{7}$ (i.e., $\frac{4}{7}(8 - 2) - 2$) which is also $A_2$'s real utility (i.e., $\frac{3}{7}(10 - 2) - 2$). $A_1$'s real utility, however, is $2\frac{4}{7} = \frac{4}{7}(10 - 2) - 2$. This lie is beneficial for $A_1$.

The situation is illustrated in Figure 29, where agent $A_1$'s lie modifies his apparent goal states so that they are closer to the initial state, but the plan still ends up bringing the world to one of his *real* goal states.

In the example above, the existence of a common subgoal between the agents allowed one agent to exploit the common subgoals (assuming, of course, that the lying agent knew its opponent's goals). The lying agent relaxes his true goal by claiming that the common subgoal is mainly its opponent's demand—as far as he is concerned (he claims), he would be satisfied with a much cheaper subgoal. If it is really necessary to achieve the expensive subgoal (he claims), more of the burden must fall on his opponent.

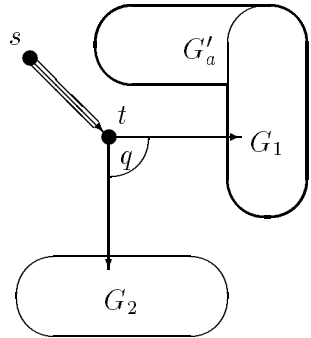

Figure 29: Lying in a Conflict Situation

One might think that in the absence of such a common subgoal, there would be no opportunity for one agent to beneficially lie to the other. This, however, is not true, as we see below.

## 7.4 Beneficial Lies with Multi-Plan Deals

**Another Example of Beneficial Lying in a Conflict Situation:** The initial state $s$ can be seen in Figure 30, similar to the example used in Section 5.5 above. $A_1$'s goal is to reverse the blocks in slot 2, and to leave the blocks in slot 1 in their initial position. $A_2$'s goal is to reverse the blocks in slot 1, and to leave the blocks in slot 2 in their initial position. To achieve his goal alone, each agent needs to do at least 8 PickUp/PutDown operations. This is a conflict situation.

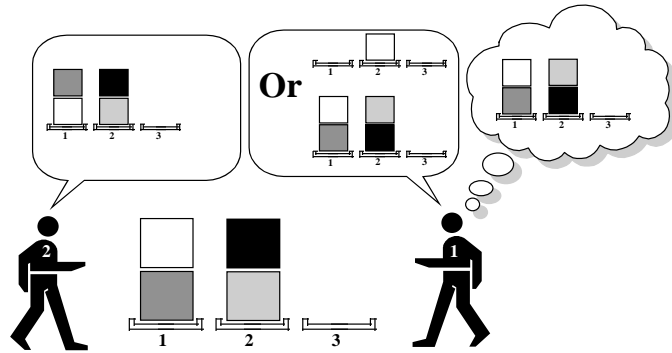

Figure 30: Example of Interference Decoy Lie

Negotiation over multi-plan deals will cause the agents to agree on $(\delta_1, \delta_2)$: $\frac{1}{2}$, where $\delta_i$ is the mixed joint plan in which both agents cooperatively achieve $i$'s goal. The best joint plan for doing the reverse in either one of the slots costs 2 PickUp/PutDown operations for each agent. Each agent's utility from this deal is $2 = (\frac{1}{2}(8 - 2) - \frac{1}{2}(2))$.

Agent $A_1$ might lie and claim that his goal is to reverse the blocks in slot 1 and leave the blocks in slot 2 in their initial position (his real goal) OR to have the white block be alone in slot 2. It costs $A_1$ 6 to achieve his apparent goal alone. To do the reverse alone would cost him 8, and thus to achieve the imaginary part of his goal is cheaper. The agreement will be $(\delta_1, \delta_2): \frac{4}{7}$, where $\delta_i$ is again the mixed joint plan in which both agents cooperatively achieve $i$'s goal. It turns out to be cheaper for both agents to cooperatively carry out $A_1$'s real goal than it is to cope with $A_1$'s imaginary alternative. $A_1$'s apparent utility will be $1\frac{3}{7} = \frac{4}{7}(6-2) - \frac{3}{7}(2)$. This is also $A_2$'s utility. $A_1$'s actual utility, however, will be $2\frac{4}{7} = \frac{4}{7}(8-2) - \frac{3}{7}(2)$, which is greater than the unvarnished utility of 2 that $A_1$ would get without lying. So even without a common subgoal, $A_1$ had a beneficial lie. Here we have been introduced to a new type of lie, a kind of "interference decoy," that can be used even when the agents' have no common subgoals.

## 8. Incomplete Information about Worth of Goals

Consider the situation where two agents encounter one another in a shared environment. Their individual goals are commonly known (because of prior knowledge about the type of agent, some goal recognition process, etc.), as well as the cost of achieving those goals, were each agent to be alone in the world. In addition, there is no conflict between these goals. There exists some non-empty set of states that satisfies both agents' goals.

The agents have upper bounds on their worth, but (in contrast to the public goals) each upper bound is private information, not known to the other agent. This is a common situation; as agents queue up to access a common resource, their goals will often be self-evident. For example, two agents approaching a narrow bridge from opposite ends may know that the other wants to cross, but not know what the crossing is worth for the other (e.g., how long it is willing to wait). The agents need to agree on a deal (for example, who will go first, and who will wait).

One simple way to design a negotiation mechanism that handles the lack of information is to have agents exchange private information prior to the actual negotiation process. This pre-negotiation exchange of information is another variant on the $-1$-phase game mentioned above. In the current case, agents exchange private information about worth. In this section, we only consider the situation where agents are negotiating over mixed joint plans.

One question, then, is how should agents play this $-1$-phase game to best advantage? As was mentioned above in Sections 4.4 and 7.2, an agent generally has an incentive to misrepresent the worth of his goal by lowering it—the less an agent is willing to pay, the less it will have to pay in a utility product maximizing mechanism (PMM). However, if everyone lowers their worth they may not be able to reach any agreement at all, whereas if they declared their true worth agreement would have been reached. Agents might lower their worth too much and be driven to an inefficient outcome. This is an instance of the free rider problem. Every agent is individually motivated to lower his worth, and have someone else carry more of the burden. The group as a whole stands to suffer, particularly if agreements are not reached when they otherwise would have been.

We can exert control over this tendency to lower one's apparent worth by careful design of the post-exchange part of the negotiation mechanism. We are interested in designing a mechanism that satisfies our desire for efficient, symmetric, simple, and stable outcomes. In our research on TODs, we managed (in certain cases) to provide a post-exchange mechanism that satisfied all these attributes, and was also found to be incentive compatible—the agents' best strategy was to declare their true goals. In this section, we introduce two mechanisms for private-worth SODs, one "strict," the other "tolerant," and analyze their affects on the stability and efficiency of negotiation outcomes. The strict mechanism turns out to be more stable, while the tolerant mechanism is more efficient.

## 8.1 Strict and Tolerant Mechanisms

There are several cases that need to be addressed by any mechanism, and can be treated differently by different mechanisms. For example, what should happen if one agent declares his worth as being lower than his stand-alone cost (i.e., apparently he would not achieve his goal were he to be alone, it is not worth it to him)? Should the other agent then still be allowed to offer a deal, or is the negotiation considered to have failed at this point?

Both mechanisms that we present below start the same way. The agents simultaneously declare a worth value, claimed to be the worth they assign to the achievement of their goal.

- **Both goals are apparently achievable alone:** If both agents declare a worth greater than their stand-alone cost (which is commonly known), the negotiation proceeds as if the worth declarations were true. The agents then use some product maximizing mechanism over the negotiation set of mixed joint plans, with their declared worths as the baseline of the utility calculations. The result will be an equal division of the apparent available utility between them.

- **Only one agent's goal is apparently achievable alone:** If one agent declares a worth greater than stand-alone cost, and the other doesn't, then the former agent is free to decide what to do. He can either propose a take-it-or-leave-it deal to the other agent (if it's refused, he'll carry out his own goal alone), or he can simply bypass the offer and just carry out his own goal. Since his declared worth is greater than his stand-alone cost, it is rational for him to accomplish his goal by himself.

- **Both agents' goals are apparently unachievable alone:** If both agents declare worths lower than their stand-alone costs, our two mechanisms differ as to how the situation is handled:

  - **Strict Mechanism:** There is a conflict, and no actions are carried out. The agents derive the utility of the conflict deal.
  - **Tolerant Mechanism:** The agents continue in their negotiation as in the first case above (i.e., they use mixed joint plans, and divide the apparent available utility between them). Even though both agents claim to be unwilling to achieve their goals alone, it may certainly be the case that together, they can carry out a rational joint plan for achieving both of their goals.

The tolerant mechanism gives the agents a "second chance" to complete the negotiation successfully and reach a rational agreement, whereas the strict mechanism does not forgive their low worth declarations, and "punishes" them both by causing a conflict. Of course, if both agents' *true* worths are really lower than their stand-alone costs, the strict mechanism causes an unnecessary failure (and is thus inefficient), while the tolerant mechanism still allows them to reach a deal when it is possible. We will see below, however, that tolerance can sometimes lead to instability.

Our approach through the rest of this section will be to consider the various relationships among the two agents' worth values, their cost values, the interaction types, and the joint plans that achieve both agents' goals. For each such relationship, we'll analyze the strategies available to the agents. As mentioned above, we are here only considering situations where both agents' goals are achievable by two-agent mixed joint plans (e.g., there are reachable states that satisfy both agents' goals).

The idea of tidy agents and an agent cleaning up after himself, introduced above in Section 6, was used in situations where agents were willing to pay any price to achieve their goals—their worths were unbounded. There, worth could not be used as a baseline for the utility calculation. Instead, we found that there was a "minimal sufficient" value to the utility baseline that gave rise to an efficient and fair mechanism. A similar idea will also be useful in our analysis below. The tidy agent baseline, explored above, also serves as a minimal sufficient declaration point when worths are private information.

## 8.2 The Variables of Interest

In general, an agent would like to declare as low a worth as possible, but without risking a conflict. The lower the declaration of worth, the smaller his share of the joint plan will be. Unfortunately for the agent, if his declared worth is *too* low, it may eliminate the possibility of reaching an agreement. A necessary and sufficient condition for the negotiation set not to be empty is that the sum and min conditions, from Section 4.1, will hold (given the declarations of worth). Since we assume that there is a joint plan that achieves both agents' goals, agreement will still be possible if among those plans there is at least one that satisfies the sum and min conditions.

There are several variables that will play a role in our analysis below. First, each agent $i$ has a stand-alone cost (known to all, and dependent only on his goal), denoted by $c_i$. Second, each agent has a true worth (privately known) that he assigns to the achievement of his goal, denoted by $w_i$. $T$ is the total cost of the minimal (total cost) joint plan that achieves both agents' goals. $M_r$ is the cost of the minimal role among all such joint plans of cost $T$. Below, we will analyze all possible configurations of these variables.

The analysis is presented according to each interaction type other than conflict, i.e., symmetric cooperative, non-symmetric cooperative/compromise, and symmetric compromise. For each type, we will consider three subcases that depend on the relationships between $c_i$, $w_i$, $T$, and $M_r$.

## 8.3 Symmetric Cooperative Situation

In symmetric cooperative situations, one strategy that an agent can use is to declare as his worth the minimum between his true worth, and the maximum of his stand-alone cost and the minimal role in the joint plan:

$$\text{Min-Sufficient Strategy} \equiv \min(w_i, \max(c_i, M_r)).$$

The motivation here is that the agent wants to declare the minimal worth sufficient for there to be an agreement. Declaring $c_i$ satisfies the sum condition, but to make sure that it also satisfies the min condition, the agent must declare $\max(c_i, M_r)$. To make sure that this declaration is individual rational, he must not make a declaration greater than his true worth, $w_i$; thus, he takes the minimum between $w_i$ and the $(c_i, M_r)$ maximum.

The Min-Sufficient Strategy is only one possible strategy that might be adopted. However, if both agents adopt it, the strategy is in equilibrium with itself (in most cases), and agreement is guaranteed. We will analyze the characteristics of this strategy below in six cases.

### 8.3.1 BOTH GOALS ARE ACHIEVABLE ALONE

In this situation (as shown in Figure 31), both agents would be able to achieve positive utility if the other agent were not around, and they achieved their stand-alone goal by themselves.

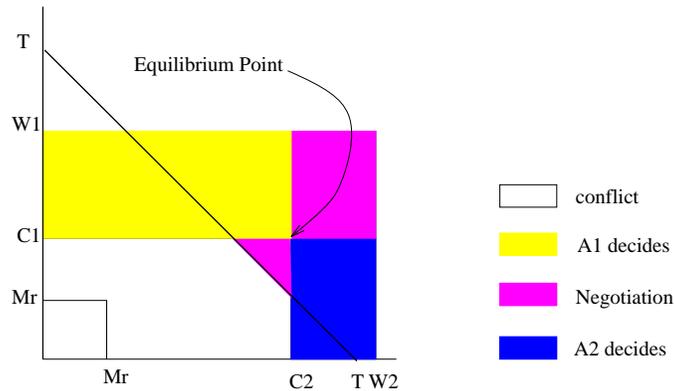

Figure 31: Both Goals are Achievable Alone

The diagram in Figure 31 describes, in a sense, the game in normal form. Each agent can declare as its worth any number from 0 to infinity. The outcome depends on the two numbers declared; every point in the plain is a possible result. The colors of the regions denote the types of outcomes.

Note, for example, that if agent $A_1$ declares less than $c_1$, while agent $A_2$ declares more than $c_2$, the outcome is that $A_2$ will decide what to do (offering $A_1$ a take-it-or-leave-it deal, or going it alone). If $A_1$ and $A_2$ both offer too little (so that the sum is less than $T$), they will reach conflict. Because we assume the agents are rational, we only consider the areas in the plain framed by $w_1$ and $w_2$ (rational agents would not declare a worth greater than their true worths).

The difference between the Strict and Tolerant mechanisms mentioned above is the color of the triangle to the lower left of the $c_1/c_2$ point. With the Strict mechanism, it would be white (conflict), while with the Tolerant mechanism it is still a region that allows subsequent negotiation to occur. The only point that is in equilibrium in both mechanisms is the $c_1/c_2$ point, which is reached by the Min-Sufficient Strategy given above. Thus, that strategy is both stable and efficient in both Strict and Tolerant mechanisms in this situation.

### 8.3.2 ONE GOAL IS ACHIEVABLE ALONE

Assume that we have the situation shown in Figure 32, where only one agent would be able to achieve positive utility if the other agent were not around (though they both ultimately can benefit from the other's existence, one more than the other).

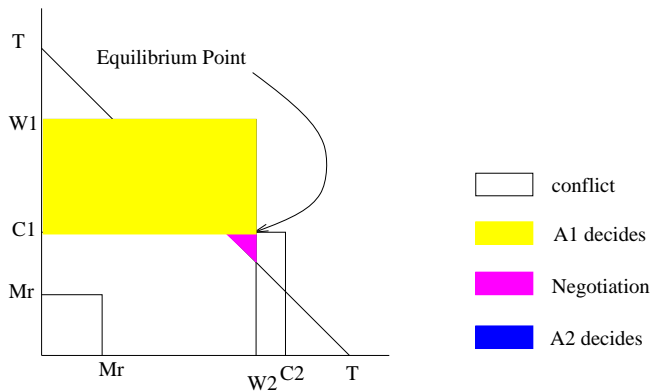

Figure 32: One Goal is Achievable Alone

We have a phenomenon similar to that in Section 8.3.1. The negotiation triangle to the lower left of $c_1/w_2$ will be white (conflict) in the Strict mechanism and negotiable in the Tolerant mechanism. In both mechanisms, the $c_1/w_2$ point is in equilibrium, which is the point that results if both agents play the Min-Sufficient Strategy. Again, that strategy is both stable and efficient in both Strict and Tolerant mechanisms in this situation.

### 8.3.3 BOTH GOALS ARE NOT ACHIEVABLE ALONE

Now consider the situation as shown in Figure 33, where neither agent could achieve positive utility were it alone in the world—the only way to achieve their goals is by cooperating.

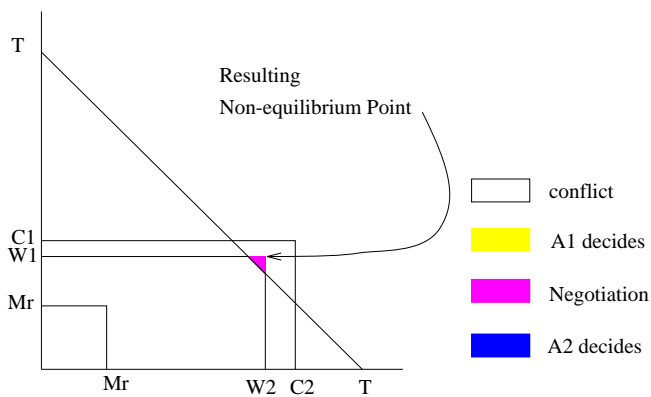

Figure 33: Both Goals are Not Achievable Alone

Again, the negotiation triangle to the lower left of $w_1/w_2$ will be white (conflict) in the Strict mechanism, and no agreement can be reached in any situation (the whole plain is, in fact, white). Though the Min-Sufficient Strategy is not efficient with the Strict mechanism, it is stable. With the Tolerant mechanism, the Min-Sufficient Strategy is efficient (it results in the $w_1/w_2$ point), but it unfortunately is not stable—assuming that one agent declares $w_1$, the other agent can benefit by declaring $T - w_1$ instead of $w_2$.

In fact, the agents do not actually know what situation they are in (the one in Figure 32 or Figure 33), so guaranteed beneficial divergence from the equilibrium point would really require total knowledge of the situation and what your opponent is playing. Thus, although the Min-Sufficient Strategy is not stable, agents may be unlikely to diverge because of real-world constraints on their knowledge.

## 8.4 Non-Symmetric Cooperative/Compromise Situation

In this section we continue our analysis into situations where for one agent, the situation is cooperative, while for the other, it is a compromise situation. We will continue to analyze the case where both agents use the Min-Sufficient Strategy. Agreement can only be reached when the compromising agent contributes more than his stand-alone cost to the joint plan; this is because the minimal role is greater than his stand-alone cost. The only way for agents to reach an agreement is when the compromising agent is willing to do more than its stand-alone cost—otherwise, there will be a conflict.

### 8.4.1 COMPROMISE IS SUFFICIENT

In the situation described by Figure 34, the true worth for the compromising agent ($w_2$) is greater than both the minimal role and $c_2$.

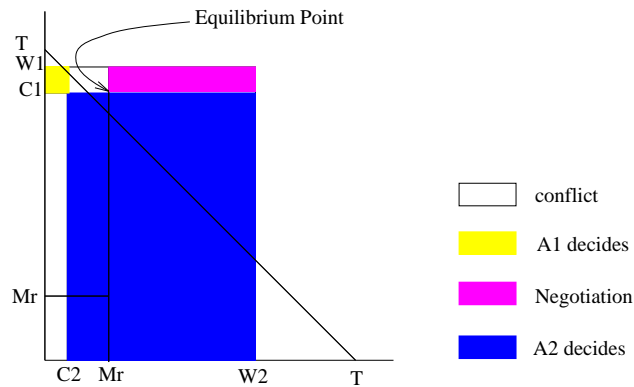

Figure 34: Compromise is Sufficient

It is sufficient for the compromising agent to declare $M_r$ as his true worth. If he declared less than that, and the other agent declared more than $c_1$, they can reach conflict; by declaring $M_r$, he guarantees that his goal will be achieved. The diagram is the same for both the Strict and Tolerant Mechanisms. The Min-Sufficient Strategy brings both agents to the $c_1/M_r$ point, which is both a stable and efficient result (for both mechanisms).

### 8.4.2 Can Compromise, But Not Enough

Consider the situation, portrayed in Figure 35, where $w_2$ is less than $M_r$, and it is not rational for agent $A_2$ to compromise and declare a worth greater than $w_2$. The Min-Sufficient Strategy brings the agents to the $c_1/w_2$ point, which is a conflict.

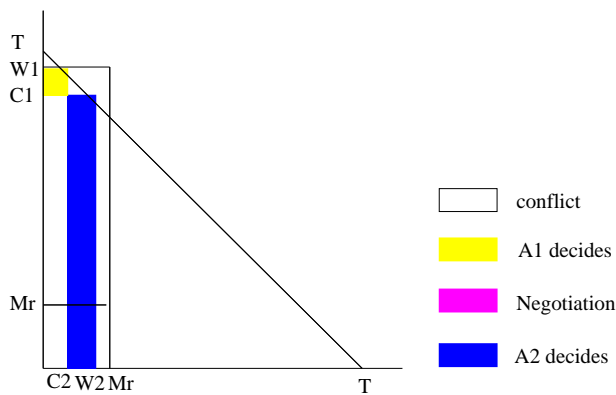

Figure 35: Can Compromise, But Not Enough

The picture is identical for both the Strict and Tolerant mechanisms. If the agents use the Min-Sufficient Strategy, the resulting point $(c_1/w_2)$ is not efficient, even though it is stable.[6] However, if we enhanced the mechanism with conflict-resolution techniques, and allowed the agents to negotiate over multi-plan deals from Section 5.5 (or even semi-cooperative deals from Section 5.3), we conjecture that the result $c_1/w_2$ would be both stable and efficient. That enhancement, however, is beyond the scope of the work described in this paper.

### 8.4.3 No Reason to Compromise

In the situation shown in Figure 36, the non-compromising agent $A_1$ cannot achieve his goal alone. The Min-Sufficient Strategy will have him declare something less than $c_1$ (either $w_1$ or $M_r$), and the result will be that agent $A_2$ will have the option to decide on what to do—and the only reasonable decision will be for $A_2$ to achieve his goal alone (there is no reason to compromise).

This result is both efficient and stable, in both the Strict and Tolerant mechanisms.

## 8.5 Symmetric Compromise Situation

In this section we continue our analysis into situations where for both agents, they are in a compromise situation. Both agents will have to do more than their stand-alone costs in order to achieve both goals.

---

6. Conflict is not efficient because the result in which one agent achieves his goal, rather than both agents doing nothing, would be more efficient.

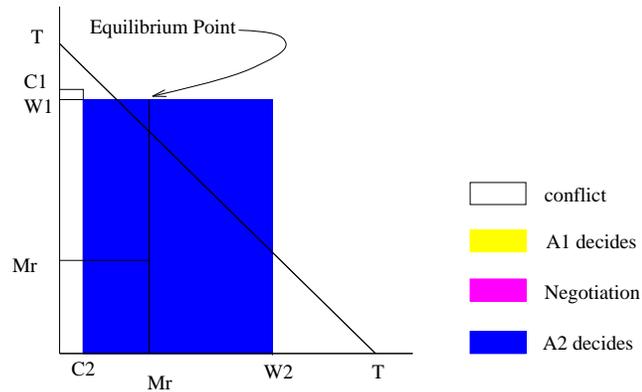

Figure 36: No Reason to Compromise

In this section, we will propose another strategy that agents could use, namely the Min-Concession Strategy:

$$\text{Min-Concession Strategy} \equiv \min(w_i, (c_i + \frac{T - (c_1 + c_2)}{2})).$$

In this situation, the agent is choosing to propose (as his true worth) more than his stand-alone cost, to ensure that an agreement can be reached. However, he would like to propose the minimal sufficient concession, just enough to enable an agreement. The Min-Concession Strategy has both agents make the same concession. The overall strategy that we are analyzing (and that covers all cases in this section) is to use the Min-Concession Strategy in symmetric compromise situations, but otherwise to use the Min-Sufficient Strategy (as presented above). Agents know which kind of situation they are in (and thus what strategy to use) because stand-alone costs are common knowledge.

### 8.5.1 AGENTS CAN COMPROMISE EQUALLY

If agents are in a situation where both can compromise equally (as shown in Figure 37), and they both use the Min-Concession Strategy, they end up at the point $(c_1 + \Delta)/(c_2 + \Delta)$ (where $\Delta = (T - c_1 - c_2)/2$). This point is both efficient and stable, under both the Strict and Tolerant mechanisms.

### 8.5.2 NON-SYMMETRIC COMPROMISE, BUT GOALS CAN BE ACHIEVED

If agents are in a symmetric compromise situation, but one in which one agent needs to compromise more than the other (as in Figure 38), the use of the Min-Concession Strategy results in the point $(c_1 + \Delta)/w_2$. This point is a conflict, and is unfortunately neither stable nor efficient.

The result is not stable because $A_1$ could make a greater compromise and benefit from it. The result is not efficient because even if only one agent could achieve its goal, that would be superior to the conflict outcome. It is not difficult to imagine other strategies that would lead the agents to efficient solutions (e.g., declare $T - c_j$ for each agent $i$, as a Tidy Agent would do in Section 6), but they would not be stable, either. On the other hand, if the negotiation mechanism is enhanced with conflict-resolution techniques (such as multi-plan deals or semi-cooperative deals), we conjecture that the Min-Concession Strategy will be both stable and efficient. This enhancement, however, is also beyond the scope of the work described in this paper.

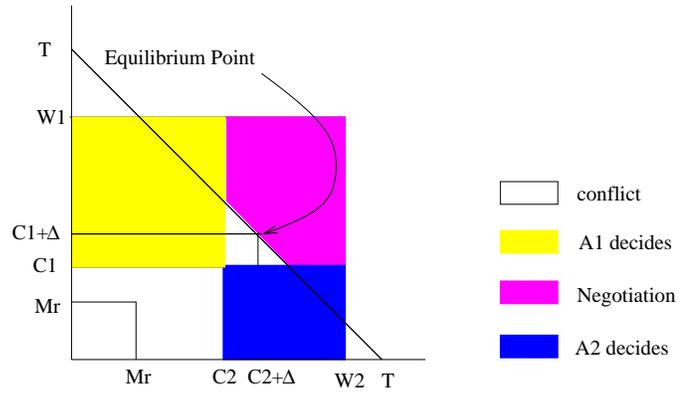

Figure 37: Agents Can Compromise Equally

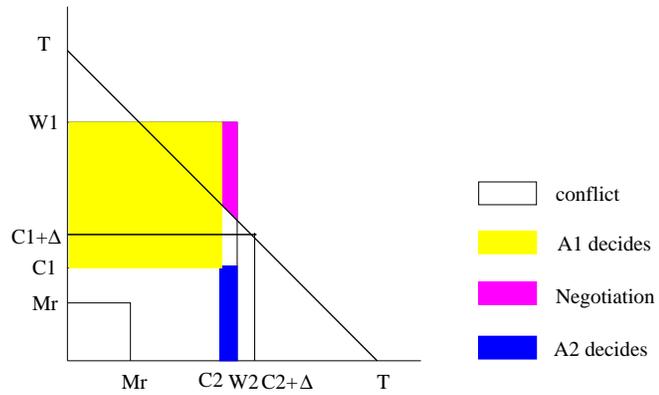

Figure 38: Non-Symmetric Compromise, but Goals can be Achieved

### 8.5.3 ONE AGENT CANNOT COMPROMISE

Consider the situation where one agent cannot compromise (because he could not even achieve his own goal alone), shown in Figure 39. In this case, if both agents use the Min-Concession Strategy, the result will be $(c_1 + \Delta)/w_2$. Agent 1 will choose to then achieve his own goal alone (and not compromise). This outcome is both stable and efficient.

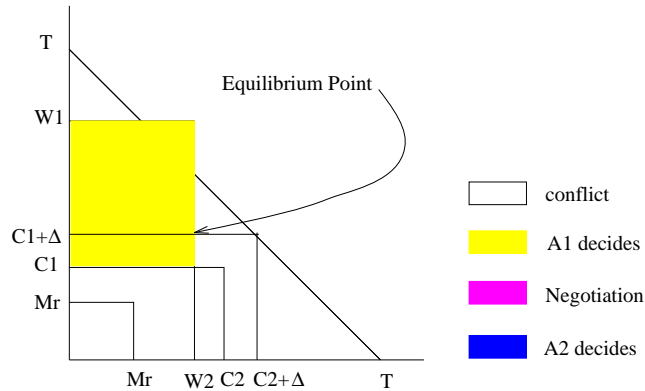

Figure 39: One Agent Cannot Compromise

## 8.6 Summary of Strict and Tolerant Mechanisms

The results of the analysis above are summarized in Figure 40. The tradeoff between efficiency and stability is apparent only in the symmetric cooperative case, where neither agent is able to achieve its goal alone. With the strict mechanism, a conflict is caused simply because each agent will not declare a worth higher than its stand-alone cost, and thus will bring about immediate conflict. The tolerant mechanism gives the agents a second chance to reach agreement, but is unstable (as described above).

The above mechanism is not incentive compatible. The agents do not have an incentive to declare their true worths; rather, they use the Min-Sufficient Strategy to decide what their optimal declaration is.

## 9. Related Work in Game Theory and DAI

In this section we review research in game theory and in distributed artificial intelligence related to our own work.

## 9.1 Related Work in Game Theory

As mentioned at the beginning of this paper, our research relies heavily on existing game theory tools that we use to design and evaluate protocols for automated agents. Here, we review the game theory work on Bargaining Theory, Mechanism Design and Implementation Theory, and Correlated Equilibria.

| | Strict | | Tolerant | |
|---|---|---|---|---|
| | **Efficient** | **Stable** | **Efficient** | **Stable** |
| **Symmetric Cooperation** | | | | |
| Both boals are achievable alone | ✓ | ✓ | ✓ | ✓ |
| One goal is achievable alone | ✓ | ✓ | ✓ | ✓ |
| Both goals aren't achievable alone | | ✓ | ✓ | |
| **Non-Symmetric Cooperation/Compromise** | | | | |
| Compromise is sufficient | ✓ | ✓ | ✓ | ✓ |
| Compromise is insufficient | | ✓ | | ✓ |
| No reason to compromise | ✓ | ✓ | ✓ | ✓ |
| **Symmetric Compromise** | | | | |
| Agents can compromise equally | ✓ | ✓ | ✓ | ✓ |
| Agents can't compromise equally | | | | |
| One agent can't compromise | ✓ | ✓ | ✓ | ✓ |

Figure 40: Summary of Strict and Tolerant Mechanisms

### 9.1.1 BARGAINING THEORY

Classic game theory (Nash, 1950; Zeuthen, 1930; Harsanyi, 1956; Roth, 1979; Luce & Raiffa, 1957) talks about players reaching "deals," which are defined as vectors of utilities (one for each player). A bargaining game can end up in some possible outcome (i.e., a "deal"). Each player has a full preference order over the set of possible outcomes; this preference order is expressed by his utility function. For each deal, there is a utility vector which is the list of the utilities of this deal for every participant. There is a special utility vector called "conflict" (or sometimes the "status quo point") which is the utility each player assigns to a conflict (that is, lack of a final agreement). Classic game theory deals with the following question: given a set of utility vectors, what will be the utility vector that the players will agree on (under particular assumptions)? In other words, classic bargaining theory is focused on prediction of outcomes, under certain assumptions about the players and the outcomes themselves.

Nash (Nash, 1950, 1953) showed that under some rational behavior assumptions (i.e., *individual rational* and *pareto optimal* behavior), and symmetry assumptions, players will reach an agreement on a deal that maximizes the product of the players' utility (see Section 4.2 for a more complete discussion).

An alternative approach to negotiation, which looks upon it as a dynamic, iterative process, is discussed in the work of Rubinstein and Osborne (Rubinstein, 1982, 1985; Osborne & Rubinstein, 1990).

Game theory work on negotiation assumes that the negotiation game itself is well-defined. It assumes that there is a set of possible deals that the players are evaluating using certain utility functions. Therefore, the deals and the players' utility functions induce a set of utility vectors that forms the basis for the negotiation game.

In contrast to this analysis of a given, well-defined negotiation encounter, we are exploring the design space of negotiation games. Given a multiagent encounter (involving, for example, task redistribution), we design an assortment of negotiation games, by formulating various sets of possible deals and various kinds of utility functions the agents may have. For any given negotiation game, we then use the above game theory approaches to analyze it and to evaluate the negotiation mechanisms we propose.

Game theorists are usually concerned with how games will be played, from both a descriptive and normative point of view. Ours is essentially a constructive point of view; since game theory tells us, for any given game, how it will be played, we endeavor to design games that have good properties when played as game theory predicts.

### 9.1.2 EQUILIBRIUM

Game solutions in game theory consist of strategies in equilibrium; if somehow a social behavior reaches an equilibrium, no agent has any incentive to diverge from that equilibrium behavior. That equilibrium is considered to be a solution to the game. There may be one or more (or no) strategies in equilibrium, and there are also different notions of equilibrium in the game theory literature.

Three levels of equilibrium that are commonly used in game theory are Nash equilibrium, perfect equilibrium, and dominant equilibrium (Binmore, 1990; Rasmusen, 1989). Each level of equilibrium enumerated above is stronger than the previous one. Two strategies $S, T$ are in Nash equilibrium if, assuming that one agent is using $S$, the other agent cannot do better by using some strategy other than $T$, and vice versa. Perfect equilibrium means that when the game has multiple steps, and one player is using $S$, there exists no state in the game where the other player can do better by not sticking to his strategy $T$. There do exist situations where strategies might be in Nash equilibrium, but not in perfect equilibrium; in that case, although strategy $T$ was best at the start of the game, as the game unfolds it would be better to diverge from $T$. Dominant strategy equilibrium means that no matter what strategy your opponent chooses, you cannot do better than play strategy $T$; strategies $S$ and $T$ are in dominant strategy equilibrium when $S$ is the dominant strategy for one player, and $T$ is the dominant strategy for the other.

In our work, we generally use Nash equilibrium (the weakest equilibrium concept) as our requirement of a solution; this provides us with the widest range of interaction solutions. At times, because the solution is not inherently in perfect equilibrium, we have introduced additional rules on the interaction, to compel agents to follow particular Nash equilibrium strategies as the game progresses (such as introducing a penalty mechanism for breaking a public commitment).

This provides an interesting example of the power we wield as designers of the game. First, we would normally require perfect equilibria in multiagent encounters, but we can adopt Nash equilibria as sufficient for our needs, then impose rules that keep agents from deviating from their Nash equilibrium strategies. Second, the very strong requirement of dominant equilibrium, which might be desirable when two arbitrary agents play a given game, is not needed when the recommended strategies are commonly known—Nash equilibrium is then sufficient.

### 9.1.3 MECHANISM DESIGN AND IMPLEMENTATION THEORY

There are also groups of game theorists who consider the problem of how to design games that have certain attributes. It is this area of *mechanism design* that is closest to our own concerns, as we design protocols for automated agents.

Mechanism design is also known in the game theory literature as the *implementation problem*. The *implementation question* (Binmore, 1992; Fudenberg & Tirole, 1992) asks whether there is a *mechanism* (also called a *game form*) with a distinguishable equilibrium point (dominant strategy, or perfect, or merely Nash) such that each social profile (i.e., group behavior) is associated, when the players follow their equilibrium strategies, with the desired outcome.

In other words, there are assumed to be a group of agents, each with its own utility function and preferences over possible social outcomes. There is also a social welfare function that rates all those possible social outcomes (e.g., a socially efficient agreement may be rated higher than a non-efficient one) (Arrow, 1963). The question is then, can one design a game such that it has a unique solution (equilibrium strategies), and such that when each individual agent behaves according to this equilibrium strategy, the social behavior will maximize the social welfare function. If such a game can be designed, then it is said that the game implements the social welfare function.

As an example of a social welfare function, consider minimization of pollution. While everyone may be interested in lowering pollution levels, everyone is interested in others bearing the associated costs. A mechanism to implement this social welfare function might include, for example, taxes on polluting industries and tax credits given for the purchase of electric cars. This is precisely the kind of mechanism that would cause agents, following an equilibrium strategy, to minimize pollution.

Given a negotiation game that we have designed (i.e., a set of deals and utility functions), we also have to design the actual negotiation mechanism. One of the important attributes of the negotiation mechanism is *efficiency*, i.e., maximization of the total group's utility. This is the *social welfare function* that we are trying to *implement*. When we assume that agents have incomplete information about one another's utility function, we basically have a (negotiation) mechanism design problem.

However, unlike classic mechanism design in game theory, we are satisfied with a (negotiation) mechanism that has *some* Nash equilibrium point that implements efficiency. We do not need uniqueness, nor do we need a stronger notion of equilibrium (i.e., dominant equilibrium). The negotiation mechanism we design is intended as a suggestion to the community of agents' designers, along with a negotiation strategy. The negotiation mechanism and the strategy are both part of the suggested *standard*. To make the standard self-enforcing it is sufficient that the strategy that is part of the standard be in Nash equilibrium.

### 9.1.4 CORRELATED EQUILIBRIUM

Players can sometime communicate prior to actually playing the game. By communicating, the players can coordinate their strategies or even sign binding contracts about the strategies they are about to use. Contracts can be of various types. An agent can commit himself to playing a pure strategy if the other agent commits to playing another pure strategy. Agents can also commit themselves to a contract in which they flip a coin and play their strategy according to the coin.

A contract can thus be seen as an agreement between the players to *correlate* their strategies. A correlated strategy in the general case is a probability distribution over all possible joint activities (i.e., strategy combinations) of the players. In order for the players to play according to some correlated strategy, there should be a mediator to conduct the lottery, choose the joint activity according to the agreed probabilities, and then suggest this strategy to the players. In some cases the mediator is assumed to release to each player information only about that player's action (strategy) in the chosen joint action, but not the other player's action.

Contracts between players can be binding; however, we cannot assume that contracts are binding in all cases. Even when contracts are not binding, some of them can be *self-enforcing*. A contract is self-enforcing if each player that signs the contract cannot do better by *not* following the contract, under the assumption that other agents *are* following the contract. If the mediator's communications are observable by all the players, then the only self-enforcing non-binding contracts are those that randomize among the Nash equilibria of the original game ((Myerson, 1991), pp. 251).

Self-enforcing contracts on correlated strategy are called *correlated equilibria*. Aumann introduced the term correlated equilibrium (Aumann, 1974); he defined the correlated equilibrium of a given game to be a Nash equilibrium of some extension of the game, where the players receive private signals before the original game is actually played. Aumann also showed (Aumann, 1987) that correlated equilibrium can be defined in terms of Bayesian rationality. Forges extended this approach to games with incomplete information (Forges, 1993).

Myerson showed that correlated equilibrium is a specific case of a more general concept of equilibrium, which he called *communication equilibrium*, in games with incomplete information (Myerson, 1982, 1991).

Some of the deal types that we have defined above involve coin flipping. This is, of course, directly related to the notion of correlated strategies. As in the correlated equilibrium theory, we also assume that agents are able to agree on deals (i.e., contracts) that involve some jointly observed random process (e.g., a coin toss). However, unlike correlated equilibrium theory, we *do* assume that contracts are binding. Therefore, we assume that agents will follow the contract (whatever the result was of the coin flip) even if it is no longer rational for the agent to do so. Relaxation of the binding agreement assumption, and designing negotiation mechanisms that are based on self-enforcing correlated strategies, are part of our future research plans.

## 9.2 Related Work in Distributed Artificial Intelligence

There have been several streams of research in Distributed Artificial Intelligence (DAI) that have approached the problem of multiagent coordination in different ways. We here briefly review some of this work, categorizing it in the general areas of multiagent planning, negotiation, social laws, and economic approaches.

### 9.2.1 MULTIAGENT PLANNING

One focus of DAI research has been that of "planning for multiple agents," which considers issues inherent in centrally directed multiagent execution. Smith's Contract Net (Smith, 1978, 1980) falls into this category, as does other DAI work (Fox, Allen, & Strohm, 1982; Rosenschein, 1982; Pednault, 1987; Katz & Rosenschein, 1993). A second focus for research has been "distributed planning," where multiple agents all participate in coordinating and deciding upon their actions (Konolige & Nilsson, 1980; Corkill, 1982; Rosenschein & Genesereth, 1985; Rosenschein, 1986; Durfee, Lesser, & Corkill, 1987; Zlotkin & Rosenschein, 1991b; Ephrati & Rosenschein, 1991; Pollack, 1992; Pope, Conry, & Mayer, 1992).

The question of whether the group activity is fashioned centrally or in a distributed manner is only one axis of comparison. Another important issue that distinguishes between various DAI research efforts is whether the goals themselves need to be adjusted, that is, whether there may be any fundamental conflicts among different agents' goals. Thus, for example, Georgeff's early work on multiagent planning assumed that there was no basic conflict among agent goals, and that coordination was all that was necessary to guarantee success (Georgeff, 1983, 1984; Stuart, 1985). Similarly, planning in the context of Lesser, Corkill, Durfee, and Decker's research (Decker & Lesser, 1992, 1993b, 1993a) often involves coordination of activities (e.g., sensor network computations) among agents who have no inherent conflict with one another (though surface conflict may exist). "Planning" here means avoidance of redundant or distracting activity, efficient exploration of the search space, etc.

Another important issue is the relationship that agents have to one another, e.g., the degree to which they are willing to compromise their goals for one another (assuming that such compromise is necessary). *Benevolent Agents* are those that, by design, are willing to accommodate one another (Rosenschein & Genesereth, 1985); they have been built to be cooperative, to share information, and to coordinate in pursuit of some (at least implicit) notion of global utility. In contrast, Multiagent System agents will cooperate only when it is in their best interests to do so (Genesereth, Ginsberg, & Rosenschein, 1986). Still another potential relationship among agents is a modified master-slave relationship, called a "supervisor-supervised" relationship, where non-absolute control is exerted by one agent over another (Ephrati & Rosenschein, 1992a, 1992b).

The synthesis, synchronization, or adjustment process for multiple agent plans thus constitute some of the (varied) foci of DAI planning research. Synchronization through conflict avoidance (Georgeff, 1983, 1984; Stuart, 1985), distribution of a single-agent planner among multiple agents (Corkill, 1979), the use of a centralized multiagent planner (Rosenschein, 1982), and the use of consensus mechanisms for aggregating subplans produced by multiple agents (Ephrati & Rosenschein, 1993b), have all been explored, as well as related issues (Cohen & Perrault, 1979; Morgenstern, 1987; von Martial, 1992a, 1992b; Kreifelts & von Martial, 1991; Kamel & Syed, 1989; Grosz & Sidner, 1990; Kinny, Ljungberg, Rao, Sonenberg, Tidhar, & Werner, 1992; Ferber & Drogoul, 1992; Kosoresow, 1993).

In this paper, we have not been dealing with the classical problems of planning research (e.g., the construction of sequences of actions to accomplish goals). Instead, we have taken as a given that the agents are capable of deriving joint plans in a domain, and then considered how they might choose from among alternative joint plans so as to satisfy potentially conflicting notions of utility. To help the agents bridge conflicts, we have introduced frameworks for plan execution (such as flipping a coin to decide which of two joint plans will be carried out), but the actual base planning mechanism is not the subject of our work.

### 9.2.2 Axiomatic Approaches to Group Activity

There exists a large and growing body of work within artificial intelligence that attempts to capture notions of rational behavior through logical axiomatization (Cohen & Levesque, 1990, 1991; Rao, Georgeff, & Sonenberg, 1991; Rao & Georgeff, 1991, 1993; Georgeff & Lansky, 1987; Georgeff, 1987; Belegrinos & Georgeff, 1991; Grosz & Kraus, 1993; Konolige, 1982; Morgenstern, 1990, 1986; Kinny & Georgeff, 1991). The approach usually centers on a formalized model of the agent's beliefs, desires, and intentions (the so-called "BDI model") (Hughes & Cresswell, 1968; Konolige, 1986). The purpose of the formal model is to characterize precisely what constitutes rational behavior, with the intent to impose such rational behavior on an automated agent. The formal axioms might be used at run-time to directly constrain an agent's decision process, or (more likely) they could be used at compile-time to produce a more efficient executable module.

The focus of this research, coming as it does from a single-agent artificial intelligence perspective, is on the architecture of a single automated agent. For example, Cohen and Levesque have explored the relationship between choice, commitment, and intention (Cohen & Levesque, 1987, 1990)—an agent should commit itself to certain plans of action, and remain loyal to these plans as long as it is appropriate (for example, when the agent discovers a plan is infeasible, the plan should be dropped).

Even when looking at multiagent systems, these researchers have examined how a member of a group should be designed—again, looking at how to design an individual agent so that it is a productive group member. For example, in certain work (Kinny et al., 1992) axioms are proposed that cause an agent, when he discovers that he will fail to fulfill his role in a joint plan, to notify the other members of his group. Axiomatizations, however, might need to deal with how *groups* of agents could have a *joint* commitment to accomplishing some goal (Cohen & Levesque, 1991), or how each agent can make interpersonal commitments without the use of such notions (Grosz & Kraus, 1993). Another use for the BDI abstractions is to allow one agent to reason about other agents, and relativize one's intentions in terms of beliefs about other agents' intentions or beliefs.

Axiomatic approaches tend to closely link definitions of behavior with internal agent architecture. Thus, the definition of commitment explored by Cohen and Levesque is intended to constrain the design of an agent, so that it will behave in a certain way. Our work, on the other hand, takes an arms-length approach to the question of constraining agents' public behavior. The rules of an encounter are really a specification of the domain (not of the agent), and an agent designer is free to build his agent internally however he sees fit. The rules themselves, however, will induce rational designers to build agents that behave in certain ways, independent of the agents' internal architectures.

### 9.2.3 SOCIAL LAWS FOR MULTIPLE AGENTS

Various researchers in Distributed Artificial Intelligence have suggested that it would be worthwhile to isolate "aspects of cooperative behavior," general rules that would cause agents to act in ways conducive to cooperation. The hypothesis is that when agents act in certain ways (e.g., share information, act in predictable ways, defer globally constraining choices), it will be easier for them to carry out effective joint action (Steeb, Cammarata, Hayes-Roth, & Wesson, 1980; Cammarata, McArthur, & Steeb, 1983; McArthur, Steeb, & Cammarata, 1982).

Moses, Shoham, and Tennenholtz (Tennenholtz & Moses, 1989; Moses & Tennenholtz, 1990; Shoham & Tennenholtz, 1992b, 1992a; Moses & Tennenholtz, 1993; Shoham & Tennenholtz, 1995), for example, have suggested applying the *society metaphor* to artificial systems so as to improve the performance of agents operating in the system. The issues that are to be dealt with when analyzing a multiagent environment concern synchronization, coordination of the agents' activities, cooperative ways to achieve tasks, and how safety and fairness constraints on the system can be guaranteed. They propose coordinating agent activity to avoid conflicts; the system will be structured so that agents will not arrive at potential conflict situations.

Thus these social laws are seen as a method to avoid the necessity for costly coordination techniques, like planning or negotiation. With agents following the appropriate social laws, the need for run-time coordination will be reduced. This is important, because although agent designers may be willing to invest a large amount of effort at design time in building effective multiagent systems, it is often critical that the run-time overhead be as low as possible.

There is a similarity between this use of pre-compiled, highly structured social laws, and our development of pre-defined interaction protocols. However, the social law approach assumes that the designers of the laws have full control over the agents; agents are assumed to follow the social laws simply because they were designed to, and not because they individually benefit from the social laws. Obeying the social laws may not be "stable"; assuming that everyone else obeys the laws, an agent might do better by breaking them. Our approach is concerned with social conventions that are stable, which will be suitable for individually motivated agents.

### 9.2.4 DECISION THEORETIC APPROACHES

There is related work in Artificial Intelligence that addresses the reasoning process of a single agent in decision-theoretic terms. In certain work (Horvitz, 1988; Horvitz, Cooper, & Heckerma, 1989; Russell & Wefald, 1989), decision-theoretic approaches are used to optimize the value of computation under uncertain and varying resource limitations. Etzioni considered using a decision-theoretic architecture, with learning capabilities, to control problem solving search (Etzioni, 1991). For an introductory treatment of decision theory itself, see Raiffa's classic text on the subject (Raiffa, 1968).

Classical decision theory research considers an agent that is "playing against nature," trying to maximize utility in uncertain circumstances. A key assumption is that "nature's" behavior is independent of the decision made by the agent. Of course, this assumption does not hold in a multiagent encounter.

The concept of "rationality," usually expressed in decision-theoretic terms, has been used to model agent activity in multiagent encounters (Rosenschein & Genesereth, 1985; Genesereth et al., 1986). Here, axioms defining different types of rationality, along with assumptions about the rationality of others, led agents to particular choices of action. In contrast to this work, our research employs standard game theory notions of equilibrium and rationality. Other discussions of the use of rationality in general reasoning can be found in Doyle's research (Doyle, 1985, 1992).

Another decision theoretic approach, taken by Gmytrasiewicz and Durfee, has been used to model multiagent interactions (Gmytrasiewicz, Durfee, & Wehe, 1991a, 1991b; Gmytrasiewicz & Durfee, 1992, 1993). It assumes no predefined protocol or structure to the interaction (in marked contrast to our research on protocol design). The research uses a decision-theoretic method for coordinating the activities of autonomous agents called the *Recursive Modeling Method*. Each agent models the other agents in a recursive manner, allowing evaluation of the expected utility attached to potential actions or communication.

### 9.2.5 ECONOMIC APPROACHES

There have been several attempts to consider market mechanisms as a way of efficiently allocating resources in a distributed system. Among the AI work is that of Smith's Contract Net (Smith, 1978, 1980; Sandholm, 1993), Malone's Enterprise system (Malone et al., 1988), and Wellman's WALRAS system (Wellman, 1992).

The Contract Net is a high-level communication protocol for a Distributed Problem Solving system. It enables the distribution of the tasks among the nodes that operate in the system. A contract between two nodes is established so that tasks can be executed; each node in the net can act either as a *manager* or as a *contractor*. A task that has been assigned to a node can be further decomposed by the contractor. A contract is established by a bidding scheme that includes the announcement of the task by the manager, and bids sent in by the potential contractors.

Enterprise (Malone et al., 1988) is a system that was built using a variation of the Contract Net protocol. The *Distributed Scheduling Protocol* locates the best available machine to perform a task. This protocol is similar to the Contract Net, but makes use of more well-defined assignment criteria.

Another system (Wellman, 1992) that takes an economic approach in solving a distributed problem through the use of a price mechanism has been explored by Wellman. Wellman uses the consumer/producer metaphor to establish a market pricing-based mechanism for task redistribution that ensures stability and efficiency. All agents act as both consumers and producers. Each distinct good has an *auction* associated with it, and agents can get the good by submitting bids in the auction for that good. The system developed by Wellman, WALRAS, computes for each market the equilibrium price.

There are two main differences between these economic approaches and our work on mechanism design. First, there is an underlying assumption in the economic approach that utility is explicitly transferable (e.g., money can be used). Our work does not involve any need for explicit utility transfer. Instead, we exploit various methods for implicit utility transfer, for example, sharing work in a joint plan, tossing a coin, etc. Of course, this constrains the available coordination mechanism, but removes an assumption (that is, the existence of money) that may not be suitable in certain multiagent environments. Second, the economic models can deal with $n$ agents in a market, while our work above deals with two-agent encounters; however, other work of ours deals with $n$-agent negotiation as a coalition formation problem (Zlotkin & Rosenschein, 1994).

### 9.2.6 Negotiation

Negotiation has been a subject of central interest in DAI, as it has been in economics and political science (Raiffa, 1982). The word has been used in a variety of ways, though in general it refers to communication processes that further coordination (Smith, 1978; Lesser & Corkill, 1981; Kuwabara & Lesser, 1989; Conry et al., 1988; Kreifelts & von Martial, 1991; Kraus, Ephrati, & Lehmann, 1991). These negotiating procedures have included the exchange of Partial Global Plans (Durfee, 1988; Durfee & Lesser, 1989), the communication of information intended to alter other agents' goals (Sycara, 1988, 1989), and the use of incremental suggestions leading to joint plans of action (Kraus & Wilkenfeld, 1991).

Interagent collaboration in Distributed Problem Solving systems has been explored in the ongoing research of Lesser, Durfee, and colleagues. Much of this work has focused on the implementation and analysis of data fusion experiments, where systems of distributed sensors absorb and interpret data, ultimately arriving at a group conclusion (Durfee & Lesser, 1987; Decker & Lesser, 1993a; Lâasri, Lâasri, & Lesser, 1990). Agents exchange partial solutions at various levels of detail to construct global solutions; much of the work has examined effective strategies for communication of data and hypotheses among agents, and in particular the kinds of relationships among nodes that can aid effective group analysis. For example, different organizations, and different methods for focusing node activity, can help the system as a whole be far more efficient.

There are two main distinctions between our work and the work of Lesser and his colleagues. First, the underlying assumption of the bulk of Lesser's work is that agents are designed and implemented as part of a unified system, and work towards a global goal. Our agents, on the other hand, are motivated to achieve individual goals. Second, unlike our formal approach to mechanism design, Lesser's work has historically been heuristic and experimental, although his more recent work has explored the theoretical basis for system-level phenomena (Decker & Lesser, 1992, 1993a, 1993b).

Sycara has examined a model of negotiation that combines case-based reasoning and optimization of multi-attribute utilities (Sycara, 1988, 1989). In particular, while we assume that agents' goals are fixed during the negotiation, Sycara is specifically interested in how agents can influence one another to *change* their goals through a process of negotiation (information transfer, etc.).

Kraus and her colleagues have explored negotiation where the negotiation time itself is an issue (Kraus & Wilkenfeld, 1991; Kraus, 1993; Kraus, Wilkenfeld, & Zlotkin, 1995). Agents may lose value from a negotiation that drags on too long, and different agents are asymmetric with regard to the cost of negotiation time. Agents' attitudes towards negotiation time directly influences the kinds of agreements they will reach. Interestingly, however, those agreements can be reached without delay. There is an avoidable inefficiency in delaying agreement. Our work, in contrast, assumes that agent utility remains constant throughout the negotiation process, and so negotiation time does not influence the agreement. Some of Kraus' work also assumes explicit utility transfer (while our work, as mentioned above, does not).

Gasser has explored the social aspects of agent knowledge and action in multiagent systems ("communities of programs") (Gasser, 1991, 1993). Social mechanisms can dynamically emerge; communities of programs can generate, modify, and codify their own local languages of interaction. Gasser's approach may be most effective when agents are interacting in unstructured domains, or in domains where their structure is continuously changing. The research we present, on the other hand, exploits a pre-designed social layer for multiagent systems.

Other work that focuses on the organizational aspects of societies of agents exists (Fox, 1981; Malone, 1986).

Ephrati and Rosenschein used the Clarke Tax voting procedure as a consensus mechanism, in essence to *avoid* the need for classical negotiation (Ephrati & Rosenschein, 1991, 1992c, 1993a). The mechanism assumes the ability to transfer utility explicitly. The Clarke Tax technique assumes (and requires) that agents are able to transfer utility out of the system (taxes that are paid by the agents). The utility that is transferred out of the system is actually wasted, and reduces the efficiency of the overall mechanism. This, however, is the price that needs to be paid to ensure stability. Again, the work we present in this paper does not assume the explicit transfer of utility. Also, the negotiation mechanism ensures stability without the inefficiency of transferring utility out of the system. However, voting mechanisms like the Clarke Tax can deal with $n$-agent agreement (not the two-agent agreement of our research), and also demonstrates a kind of dominant equilibrium (in contrast to our weaker notion of Nash equilibrium).

## 10. Conclusions

In this paper we have explored State Oriented Domains (SODs). In State Oriented Domains the current description of the world is modeled as a state, and operators cause the world to move from one state to another. The goal of an agent is to transform the world into one of some collection of target states. In SODs, real conflict is possible between agents, and in general, agents may find themselves in four possible types of interactions, *symmetric cooperative*, *symmetric compromise*, *non-symmetric cooperative/compromise*, and *conflict*. Agents can negotiate over different deal types in each of these kinds of interactions; in particular, we introduced the semi-cooperative deal, and multi-plan deals, for use in conflict situations. The Unified Negotiation Protocols, product maximizing mechanisms based on either semi-cooperative deals or multi-plan deals, provide a suitable basis for conflict resolution, as well as for reaching cooperative agreements.

Strategic manipulation is possible in SODs. In a State Oriented Domain, an agent might misrepresent his goals, or his worth function, to gain an advantage in a negotiation. The general approach by a deceitful agent would be to pretend that its worth is lower than it actually is. This can be done directly, by declaring low worth (in certain mechanisms), or by declaring a cheaper goal (in the case where stand-alone cost is taken to be the implicit worth baseline). We were able to construct incentive compatible mechanisms to be used when worths are unknown, but were unable to do so for SODs when goals are unknown.

## Acknowledgements

This paper was submitted while Gilad Zlotkin was affiliated with the Center for Coordination Science, Sloan School of Management, MIT. This research began while Zlotkin was affiliated with the Institute of Computer Science at the Hebrew University of Jerusalem, and was supported by the Leibniz Center for Research in Computer Science. Some material in this paper has appeared in preliminary form in AAAI, IJCAI, and ICICIS conference papers (Zlotkin & Rosenschein, 1990, 1991b; Rosenschein, 1993; Zlotkin & Rosenschein, 1993c) and in a journal article (Zlotkin & Rosenschein, 1991a) (earlier version of material on the UNP protocol). This research has been partially supported by the Israeli Ministry of Science and Technology (Grant 032-8284) and by the Israel Science Foundation (Grant 032-7517). We would like to thank the anonymous reviewers who contributed to the improvement of this paper.